\documentclass[lettersize,journal]{IEEEtran}
\usepackage{amsmath,amsfonts}
\usepackage{algorithm}
\usepackage{algpseudocode}
\usepackage{array}
\usepackage[caption=false,font=normalsize,labelfont=sf,textfont=sf]{subfig}
\usepackage{textcomp}
\usepackage{stfloats}
\usepackage{url}
\usepackage{verbatim}
\usepackage{graphicx}

\usepackage {adjustbox}
\usepackage{booktabs}
\usepackage{enumitem}
\usepackage{multirow}
\usepackage{colortbl}

\newtheorem{theorem}{Theorem}

\newtheorem{theorem*}{Theorem}

\def\BibTeX{{\rm B\kern-.05em{\sc i\kern-.025em b}\kern-.08em
    T\kern-.1667em\lower.7ex\hbox{E}\kern-.125emX}}
\usepackage{balance}

\usepackage{xr-hyper}
\begin{document}
\title{Multi-Level Contextual Token Relation Modeling for Machine-Generated Text Detection}

\author{Chenwang~Wu,
        Yiu-ming~Cheung,~\IEEEmembership{Fellow,~IEEE,}
        Bo~Han,~\IEEEmembership{Senior Member,~IEEE,}
        Shuhai~Zhang,
        and Defu Lian
\IEEEcompsocitemizethanks{\IEEEcompsocthanksitem Chenwang Wu, Yiu-ming Cheung, and Bo Han are affiliated with the Department of Computer Science, Hong Kong Baptist University, Hong Kong, China.\protect\\
E-mail: \{cscwwu, ymc, bhanml\}@comp.hkbu.edu.hk.
\IEEEcompsocthanksitem Shuhai Zhang is affiliated with the School of Software Engineering, South China University of Technology, Guangzhou, Guangdong 510000, China.\protect\\
E-mail: shuhaizhangshz@gmail.com.
\IEEEcompsocthanksitem Defu Lian is affiliated with the School of Computer Science and Technology, University of Science and Technology of China, Hefei, Anhui
230000, China. He is also affiliated with the State Key Laboratory of Cognitive Intelligence.\protect\\
E-mail: liandefu@ustc.edu.cn.
\IEEEcompsocthanksitem Corresponding author: Yiu-ming~Cheung.}
}

\markboth{Multi-Level Contextual Token Relation Modeling for Machine-Generated Text Detection}%
{How to Use the IEEEtran \LaTeX \ Templates}

\maketitle

\begin{abstract}
Machine-generated texts (MGTs) pose risks such as disinformation and phishing, underscoring the need for reliable detection. Metric-based methods, which extract statistically distinguishable features of MGTs, are often more practical than complex model-based methods that are prone to overfitting. Given their diverse designs, we first place representative metric-based methods within a unified framework, enabling a clear assessment of their advantages and limitations. Our analysis identifies a core challenge across these methods: the token-level detection score is easily biased by the inherent randomness of the MGTs generation process. Then, we theoretically derive the multi-hop transitions of the token-level detection score and explore their local and global relations. Based on these findings, we propose a multi-level contextual token relation modeling framework for MGT detection. Specifically, for local relations, we model them through a lightweight Markov-informed calibration module that refines token-level evidence before aggregation. For global relations, we introduce a rule-support reasoning module that uses explicit logical rules derived from contextual score statistics. Finally, we combine the local calibrated score and the global rule-support reasoning signal in a joint multi-level inference framework. Extensive experiments show broad and substantial improvements across various real-world scenarios, including cross-LLM and cross-domain settings, with low computational overhead.

\end{abstract}


\section{Introduction}
\IEEEPARstart{G}{enerative} AI, represented by large language models (LLMs) \cite{achiam2023gpt,radford2019language}, has been advancing rapidly, and the machine-generated texts (MGTs) they produce often match human writing in fluency, coherence, and diversity. While this technological breakthrough offers immense opportunities, it has also triggered widespread societal concerns, including the spread of disinformation \cite{vykopal2024disinformation}, violations of intellectual property rights \cite{yu2023codeipprompt}, and phishing attacks \cite{hong2012state}. Therefore, the research and development of MGT detection technologies hold significant theoretical and practical value in uncovering the distinct patterns of generated text and ensuring a trustworthy AI environment.

An effective detection method is to identify LLM watermarks \cite{hou2024semstamp}, but this requires injecting watermarks into the LLM, which is often impractical due to high access permissions. Therefore, passive detection methods, including model- and metric-based approaches, have attracted significant attention. Model-based methods use a set of human- and machine-generated texts to train a binary classifier, such as OpenAI detector \cite{solaiman2019release}, ChatGPT detector \cite{guo2023close}, SeqXGPT \cite{wang2023seqxgpt}, and CoCo \cite{liu2022coco}. However, such models are often too complex, leading to overfitting to the training data. Instead, metric-based methods exploit the inherent statistical biases of LLMs to discriminate MGTs, which are model-agnostic and have better generalization properties. These methods use metrics such as log-likelihood, log-rank, and entropy. Furthermore, methods such as DetectGPT \cite{mitchell2023detectgpt}, FastDetectGPT \cite{bao2024fast}, and Binoculars \cite{hans2024spotting} detect MGTs by comparing the differences between a given text and a perturbed, regenerated, or continued text from an alternative model.

Despite their diverse designs, this paper first systematically examines several representative approaches, including Log-Likelihood \cite{solaiman2019release}, Entropy \cite{gehrmann2019gltr}, Binoculars \cite{guo2024biscope}, DetectGPT \cite{mitchell2023detectgpt}, FastDetectGPT \cite{bao2024fast}, and DNA-DetectLLM \cite{zhudna}, and situates them within a unified framework, thereby revealing their commonalities: They first compute token-level detection scores, and then employ various carefully designed strategies to aggregate them into text-level scores to make threshold-based decisions. This unified view reveals a common challenge across the existing methods: The token-level score is easily biased by the inherent randomness of the LLM generation process, while subsequent aggregation steps fail to correct for the underlying imprecision. As a result, the detection performance is tightly constrained by the precision of token-level scores. Given that token-level scores are tied to the generation process and context-dependent, a natural question arises: can we explicitly reveal and exploit contextual relationships among token-level detection scores to improve detection?

In our preliminary work \cite{wubeyond}, we attempted to answer this question from a local perspective. Starting from a theoretical bound on attention-score evolution in a simplified transformer, we were led to two important findings regarding local contextual relations: \emph{Neighbor Similarity}, namely, adjacent tokens tend to exhibit similar detection scores, and \emph{Initial Instability}, namely, early-position token scores are more unstable than later ones. Building on these two observations, we proposed a Markov-informed score calibration method that models local contextual dependence through a pairwise Markov random field and implements it efficiently via a mean-field approximation. This calibration module can be stacked on top of existing detectors to refine token-level scores before final aggregation, thereby improving detection with negligible computational overhead.

Nevertheless, purely local contextual modeling is still insufficient. While it can correct short-range score bias, it cannot adequately capture the global organization of the token-level score across the text. In the paper, we therefore go beyond the local contextual relation and propose multi-level contextual token relation modeling for MGT detection. Specifically, we extend theoretical results from attention scores to more direct token scores and from single-hop local transitions to multi-hop contextual relations. It reveals that the score differences between distant positions are also structurally bounded rather than arbitrary. This implies {global contextual relations}: within non-initial text segments, token scores exhibit global relationships, including \emph{score stability}, \emph{adjacent-difference stability}, and \emph{long-range stability} for MGT. Furthermore, this paper introduces a rule-support reasoning module to model these relations. Specifically, based on these relations, we extract corresponding global statistics from the text and construct logical rules; we then employ rule-support reasoning to derive a confidence score, which complements the locally calibrated score from our preliminary work. In this way, local Markov calibration improves the quality of token-level scores, whereas the global rule-support reasoning module provides a confidence score to enhance detection, together forming a unified multi-level contextual token relation modeling framework for MGT detection.

Except for contributions in the preliminary work \cite{wubeyond} on score calibration, this paper further makes the following contributions.
\begin{itemize}[leftmargin=*]
    \item We extend the theoretical analysis from attention scores to more direct token scores and from single-hop to more general multi-hop transitions, thereby revealing global contextual relations among token-level scores.
    \item We propose a rule-support reasoning module to explicitly capture these global relations by constructing logical rules from global statistics.
    \item We propose a multi-level framework for modeling contextual token relations to enhance MGT detection by integrating local Markov calibration and global rule-support reasoning.
    \item Extensive experiments demonstrate the effectiveness of the proposed approach in various real-world scenarios, including cross-LLM generalization, cross-domain transfer, adversarial/paraphrasing settings, and mixed text detection.
\end{itemize}

\section{Related Work}

This section provides an overview of the existing detection methods, which can be categorized into active watermark-based methods and passive model- and metric-based methods.

\subsection{Watermark-based Detection}

Watermarking is a proactive defense technique that embeds verifiable information during text generation, thereby enabling simple and reliable detection. RedList \cite{kirchenbauer2023watermark} is a model-agnostic watermarking method that dynamically partitions the vocabulary into a “greenlist” and “redlist” based on preceding context, slightly increasing the probability of sampling tokens from the greenlist. Subsequent works have made various improvements to this approach. For instance, SemStamp \cite{hou2024semstamp} introduces a sentence-level semantic hashing watermark to enhance robustness against paraphrasing attacks; DiPmark \cite{wu2023dipmark} designs an unbiased watermark that does not alter the original output distribution. REMARK-LLM \cite{zhang2024remark} is a training-based watermarking method that employs a message encoding module to generate an encrypted token distribution for watermark embedding prior to inference. Beyond manually designed watermarks, directly leveraging language models to learn to generate watermarked text is also promising \cite{liuadaptive}.

\subsection{Model-based Detection}
Model-based methods represent a classical paradigm in detection, training a binary classifier on a dataset containing both human- and machine-generated texts. A series of works, such as OpenAI Detector \cite{solaiman2019release}, ChatGPT Detector \cite{guo2023close}, GPTZero \cite{gptzero}, and G3 Detector \cite{zhan2023g3detector}, collect texts generated by various LLMs to train a unified classifier. GPT-Pat \cite{yu2023gpt} finds that detectors trained solely on a single decoding strategy generalize poorly, thereby enhancing performance by utilizing mixed decoding strategies. In addition to original data, GLTR \cite{gehrmann2019gltr} trains a simple logistic regression classifier by analyzing the predicted ranking of each word within its context. SeqXGPT \cite{wang2023seqxgpt} treats the sequence of logits as waveform signals for detection. Beyond the data level, recent works have explored more advanced training strategies. For example, LLMDet \cite{wu2023llmdet} leverages the perplexity of surrogate models as additional features; MPU \cite{tianmultiscale} adopts a positive-unlabeled learning paradigm; and RADAR \cite{hu2023radar} incorporates adversarial training to enhance model robustness. The above methods generally assume a known text source, but when the source is unknown, Ghostbuster \cite{verma2024ghostbuster} proposes training classifiers directly on texts generated by known surrogate models. Besides, DGM$^4$ \cite{shao2024detecting} incorporated contrastive learning with the image modality to capture more fine-grained data features.

\begin{table*}[t]
\centering
\caption{Comparing existing metric-based methods from a unified view. Here, $s$ is the text to be detected containing $N$ tokens, $s'$ is the perturbed text generated by DetectGPT, $\tilde{s}$ is the regenerated text of Fast-DetectGPT, and $s^*$ is the ideal text of DNA-DetectLLM. Function $\mu(\cdot)$ and $\sigma(\cdot)$ represent the mean and standard deviation of the given set, respectively. For methods whose original score directions differ, we apply sign normalization so that a larger score consistently indicates a higher likelihood of MGT.}
\label{tab: unified_view}
\renewcommand\arraystretch{0.9}
\begin{adjustbox}{width=1.\textwidth}
\begin{tabular}{c|c|c|c|c}
\toprule
\textbf{Method} & \textbf{Data} & \textbf{Token-level Score $d(s_t)$} & \textbf{Score Aggregation} & \textbf{Detection} \\
\midrule
Log-likelihood & $s$ & $\log p(s_t|s_{<t})$ & $\frac{1}{N-1}\sum_{t=2}^{N} \log p(s_t|s_{<t})$ & $score > \epsilon$ \\
Entropy & $s$ & $\sum_{v \in V} p(v|s_{<t}) \log p(v|s_{<t})$ & $\frac{1}{N-1}\sum_{t=2}^{N} \sum_{v \in V} p(v|s_{<t}) \log p(v|s_{<t})$ & $score > \epsilon$ \\
Binoculars & $s$ & $p(s_t|s_{<t})$, $q(s_t|s_{<t})$  & $\frac{-\frac{1}{N-1}\sum_{t=2}^{N} \sum_{v \in V} p(v|s_{<t}) \log p(v|s_{<t})}{-\frac{1}{N-1}\sum_{t=2}^{N} \sum_{v \in V} p(v|s_{<t}) \log q(v|s_{<t})}$ & $score > \epsilon$ \\
DetectGPT & $\{s, s'_1, s'_2, ..., s'_n\}$ & $p(s_t|s_{<t})$ &$\frac{\frac{1}{N-1} \sum_{t=2}^{N} p(s_t|s_{<t})-\mu\left(\{\frac{1}{N-1} \sum_{t=2}^{N} p(s'_{i,t}|s'_{i,<t})\}_i\right)}{\sigma\left(\{\frac{1}{N-1} \sum_{t=2}^{N} p(s'_{i,t}|s'_{i,<t})\}_i\right)}$ & $score > \epsilon$ \\
Fast-DetectGPT & $\{s, \tilde{s}_1, ..., \tilde{s}_n\}$ & $p(s_t|s_{<t})$  & $\frac{\frac{1}{N-1} \sum_{t=2}^{N} p(s_t|s_{<t})-\mu\left(\{\frac{1}{N-1} \sum_{t=2}^{N} p(\tilde{s}_{i,t}|s,\tilde{s}_{i,<t})\}_i\right)}{\sigma\left(\{\frac{1}{N-1} \sum_{t=2}^{N} p(\tilde{s}_{i,t}|s,\tilde{s}_{i,<t})\}_i\right)}$ & $score > \epsilon$ \\
DNA-DetectLLM & $\{s,s^*\}$ &
$p(s_t\mid s_{<t})$, $q(s_t\mid s_{<t})$,  $p(s^*_t\mid s_{<t})$&
$\frac{-\frac{1}{N} \sum_{t=1}^N \log p\left(s^*_t \mid s_{<t}\right)-\frac{1}{N} \sum_{i=1}^N \log p\left({s}_t \mid s_{<t}\right)}{\frac{2}{N} \sum_{t=1}^N p\left(s_t \mid s_{<t}\right) \log q\left(s_t \mid s_{<t}\right)}$. &$score> \epsilon$\\
\bottomrule
\end{tabular}
\end{adjustbox}
\end{table*}

\subsection{Metric-based Detection}

Metric-based methods do not require training on specific datasets; instead, they directly leverage the inherent statistical biases or intrinsic properties of language model-generated text to distinguish it. Early studies mainly relied on token-level probability statistics, such as Log-Likelihood \cite{solaiman2019release}, Log-Rank \cite{mitchell2023detectgpt}, and Entropy \cite{gehrmann2019gltr}, and their variants \cite{su2023detectllm,zhouadadetectgpt}. Beyond these direct scoring methods, perturbation- or rewrite-based approaches detect machine-generated text by comparing the original text with perturbed, continued, or rewritten variants, including DetectGPT \cite{mitchell2023detectgpt}, Fast-DetectGPT \cite{bao2024fast}, DNA-GPT \cite{yangdna2024}, DetectGPT4Code \cite{yang2023zero}, SimLLM \cite{nguyen2024simllm}, and L2D \cite{zhou2026learn}.
A growing line of work explores deeper intrinsic signals in text representations. These include intrinsic dimensionality \cite{tulchinskii2024intrinsic}, token coherence \cite{ma2024zero}, vocabulary-space distribution gaps \cite{yu2024text}, surrogate-model activation features \cite{chen2025repreguard}, temporal patterns of token probabilities \cite{xutraining}, relative probability spectra \cite{xu2024detecting}, and uncertainty in style perception \cite{wu2025moses}. More recent methods further model higher-level structure and robustness: DETree captures hierarchical clustering relations in hybrid human--AI text \cite{hedetree}, DNA-DetectLLM measures the repair effort needed to transform text into an ideal machine-generated sequence \cite{zhudna}, OOD-based methods improve generalization by framing human text as out-of-distribution \cite{zenghuman}, IPAD enhances interpretability by inferring likely prompts \cite{chen2025ipad}, and HLD-Detector models human and machine linguistic distributions across lexical, syntactic, and semantic levels \cite{guohld}.

\section{A Unified Perspective of Metric-based Detection}

Although model-based methods have shown competitive potential in specific domains, they are often too complex, leading to a tendency to overfit their training data. This limitation hinders their generalizability. In contrast, metric-based methods extract discriminative features from MGT, and their model-agnostic nature provides superior generalization potential. Given the diverse implementations of representative metric-based methods such as Log-Likelihood \cite{solaiman2019release}, Entropy \cite{gehrmann2019gltr}, Binoculars \cite{guo2024biscope}, DetectGPT \cite{mitchell2023detectgpt}, FastDetectGPT \cite{bao2024fast}, and DNA-DetectLLM \cite{zhudna}, we first provide a systematic examination of them from a unified perspective. This facilitates a deeper understanding of their mechanisms and allows for a fair comparison between them. As summarized in Table \ref{tab: unified_view}, we compare these methods across data, score aggregation, and detection dimensions. Note that we do not discuss their diverse core metric designs here, because our purpose is to identify common structural properties that may support a detector-agnostic enhancement framework.

\begin{itemize}[leftmargin=*]
    \item \textbf{Data}. Some methods, such as Log-Likelihood and Entropy, operate directly on the input text $s$ and are therefore computationally efficient. However, the randomness inherent in the LLM sampling mechanism may cause the MGT to deviate from these methods' underlying assumptions, e.g., Log-likelihood assumes that the generated tokens have a high likelihood. This makes it difficult for methods relying on single samples to fully exploit the potential of their core mechanisms. In contrast, DetectGPT, Fast-DetectGPT, and DNA-DetectLLM incorporate multiple perturbed (i.e., $s'$) or regenerated (i.e., $\tilde{s}$ and $\hat{s}$) samples, which mitigates the errors caused by randomness. However, this may increase computational overhead compared to single-text-based methods.
    \item \textbf{Score Aggregation}. Although these methods appear to calculate scores differently, they all tend to directly aggregate token scores to obtain the final text score. As discussed, the randomness introduced by the LLM generation process may bias token-level scores. Therefore, aggregating these potentially imprecise token scores directly may not fully reflect their core detection advantages.
    \item \textbf{Detection}. These methods employ threshold-based detection mechanisms, whose effectiveness relies heavily on the accuracy of their calculated scores. Including uncalibrated, high-noise scores in threshold-based decision-making may lead to poor performance.
\end{itemize}

In summary, existing metric-based methods improve detection in various ways, e.g., by introducing auxiliary text or redesigning scoring functions. However, they fail to address the underlying token-level errors arising from inherent randomness, thereby limiting their detection potential. Considering that detection scores are tied to tokens and LLMs' generative mechanisms induce dependencies among tokens, revealing and modeling contextual relations among token scores may help correct score errors and thus improve detection effectiveness.

\section{Contextual Relations among Token-level Scores in MGT Detection}

To understand the relationship between context tokens' detection scores, we follow existing work \cite{liu2023scissorhands} and consider the token generation process of a simplified single-layer transformer model with single-head attention\footnote{The theoretical framework is not intended to precisely characterize the full-scale LLM, but rather to reveal potential contextual relations inherent in token-level detection scores. Compared to the intractable multi-layer, multi-head Transformer, the single-layer, single-head setting serves as an analytically tractable surrogate, and our empirical observations (i.e., Figs. \ref{fig:motivation_relation_new_Essay_loss}, \ref{fig:motivation_position_new_Essay_loss}, and \ref{fig: motivation_position_DetectLLM_Essay}) further support such relations.}:
\begin{equation}
\begin{gathered}
x_{t+1}=\mathcal{F}\left(\alpha_tX_{t-1}W_VW_O\right), \text { where } \\
\alpha_t=\operatorname{softmax}\left(1 / t \cdot x_t W_Q W_K^{\top} X_{t-1}^{\top}\right).
\end{gathered}
\label{eq: simplified_model}
\end{equation}
Here, $\alpha_t$ denotes the attention scores. $x_t$ is the embedding of token $s_t$. The matrix $X_{t-1}$ is stacked by the embeddings $x_1, \ldots, x_{t-1}$, where the $j$-th row is $x_j$. $W_Q, W_K, W_V$ and $W_O$ are the attention parameters. Following the attention block, an MLP block, denoted as $\mathcal{F}(\cdot)$, is applied, and it is a two-layer network with skip connections:
$$
\mathcal{F}(x)=x+W_2 \operatorname{relu}\left(W_1 x\right).
$$

Based on this definition, we can derive the following local and global relations regarding the context detection score.

\subsection{Local Contextual Relations}
\label{sec: local}
The following result characterizes how token-level detection scores vary between adjacent positions.
\begin{theorem}
Let $d(s_t)=\mathcal{G}(\alpha_t)$ denote the token-level detection score at the $t$-th step, where $\mathcal{G}$ is $L$-Lipschitz with respect to the $L_{\infty}$-norm. Let $\lambda_K, \lambda_Q, \lambda_V, \lambda_O$ be the largest singular values of parameters $W_K, W_Q, W_V, W_O$, respectively, and let $W=W_V W_O W_Q W_K^{\top}$. For the transformer defined in Eq. (\ref{eq: simplified_model}), assuming normalized inputs ($\left\|x_t\right\|_2=1$ for all $t$) and constants $c, \epsilon>0$, consider $a_t x_{t+1}^{\top} \geq(1-\delta)\left\|a_t\right\|_2$ with $\delta \leq\left(\frac{c \epsilon}{\lambda_Q \lambda_K \lambda_V \lambda_O}\right)^2$, where $a_t=\alpha_tX_{t-1}W_VW_O$. If $x_{\ell}$ satisfies $x_{\ell} W x_{\ell}^{\top} \geq c$ and $x_{\ell} W x_{\ell} \geq \epsilon^{-1} \max _{j \in[\ell], j \neq \ell} x_j W x_{\ell}^{\top}$, then
\begin{equation}
\left|d(s_{t+1})-d(s_t)\right| \leq \frac{L(1+\sqrt{2}) \epsilon \cdot \max _j\left(x_j W x_j^T\right)}{(t+1)\left\|a_t\right\|_2}
\end{equation}
\label{theorem: local_relation}
\end{theorem}

The proof is in Section I of the Supplementary Material \footnote{\url{https://github.com/Daftstone/Multi-level-MGT-Detection}}. Theorem \ref{theorem: local_relation} characterizes how token-level detection scores evolve across local positions by directly bounding the score difference $|d(s_{t+1})-d(s_t)|$. Therefore, adjacent token scores are not independent; instead, they are locally coupled through the autoregressive generation process, and this reveals two key properties:
\begin{itemize}[leftmargin=*]
    \item \textbf{Neighbor Similarity}. A direct implication of Theorem \ref{theorem: local_relation} is that token-level evidence is expected to evolve smoothly over short contextual ranges rather than fluctuate arbitrarily from one step to the next, since their score difference is explicitly bounded. To empirically validate this property, we evaluated the distance (mean absolute difference) in detection scores across k hops (i.e., $\frac{1}{|S|}\sum_{s\in S}\frac{1}{N-k}
\sum_{t=1}^{N-k}|d(s_{t+k})-d(s_t)|$, where $S$ is the text set, and $d(s_t)$ is provided in Table \ref{tab: unified_view}). As illustrated in Fig. \ref{fig:motivation_relation_new_Essay_loss} (more results can be found in Section V-A of the Supplementary Material), the score distance increases consistently with hop size, and adjacent tokens always exhibit the strongest similarity, thereby providing empirical evidence for this neighbor similarity property.
    \item \textbf{Initial Instability}. This theorem also suggests that the detection scores of initial tokens are statistically less stable than those of subsequent tokens. This is because the bound depends on the current step $t$, it is looser at the beginning of a sequence. To validate the initial instability property, we analyzed the distance in detection score between adjacent tokens at position $t$ (i.e., $\frac{1}{|S|}\sum_{s\in S}|d(s_{t+1})-d(s_{t})|$). As shown in Fig. \ref{fig:motivation_position_new_Essay_loss} (more results are provided in Section V-B of the Supplementary Material), the adjacent score difference is much larger near the beginning of the text and then gradually decreases before stabilizing. Combined with the neighbor-similarity result, this confirms that early token-level evidence is substantially less reliable.
\end{itemize}

\begin{figure}[t]
	\centering
	\includegraphics[width=0.95\linewidth]{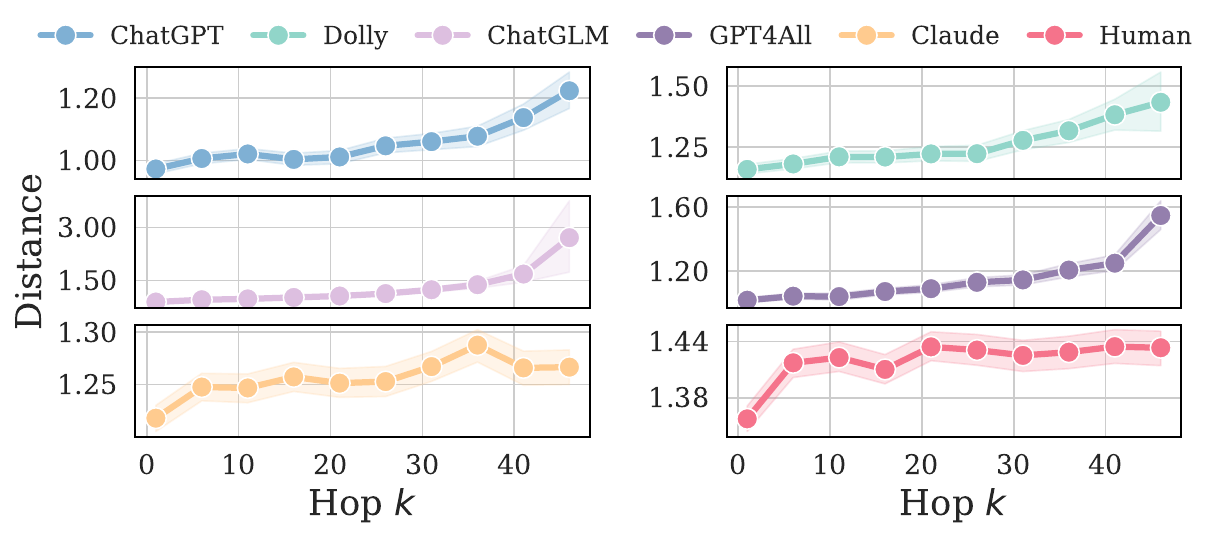}
 \vspace{-0.1cm}
	\caption{The detection score distances (Mean Absolute Difference) of neighbors with different hops in the Essay dataset. Log-likelihood scores are used.}
	\label{fig:motivation_relation_new_Essay_loss}
    \vspace{-0.2cm}
\end{figure}

\begin{figure}[t]
	\centering
	\includegraphics[width=0.95\linewidth]{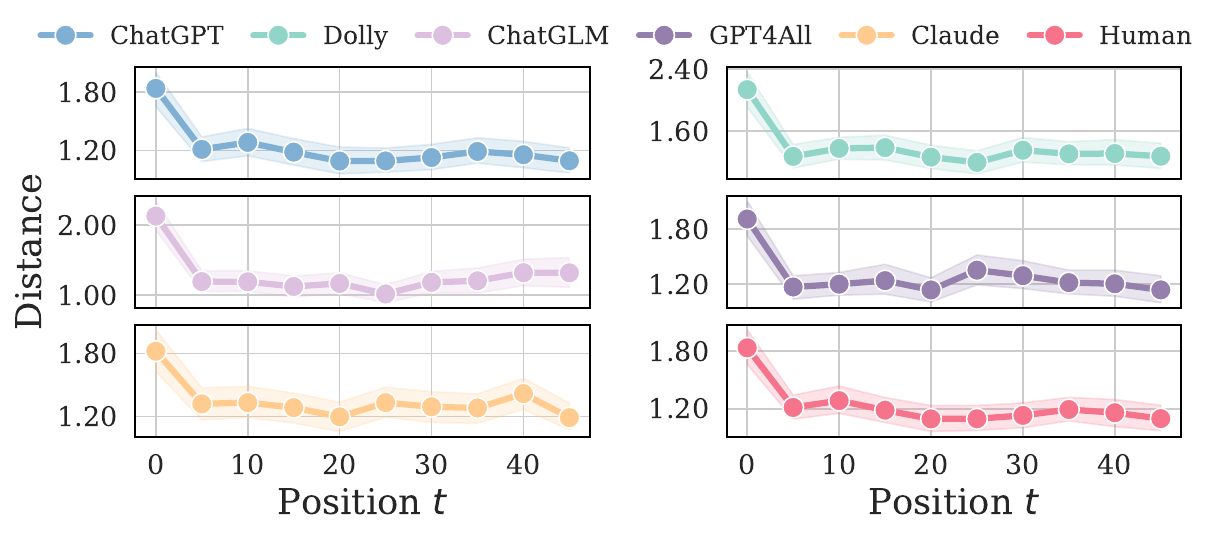}
 \vspace{-0.1cm}
	\caption{The detection score distances (Mean Absolute Difference) of 1-hop neighbors at different positions in Essay. Log-likelihood and Log-Rank score are used.}
	\label{fig:motivation_position_new_Essay_loss}
    \vspace{-0.2cm}
\end{figure}

These results reveal a clear local structure of token-level evidence: nearby scores tend to be smooth, but this smoothness is significantly weaker at early positions. Notably, although theoretical results impose no constraints on human texts, our empirical results suggest that they exhibit similar relations, given that LLMs mimic human text. This motivates the local relation modeling introduced in Section \ref{sec: local_modeling}.

\subsection{Global Contextual Relations}
\label{sec: global}

\begin{figure*}[t]
	\centering
	\includegraphics[width=1.\linewidth]{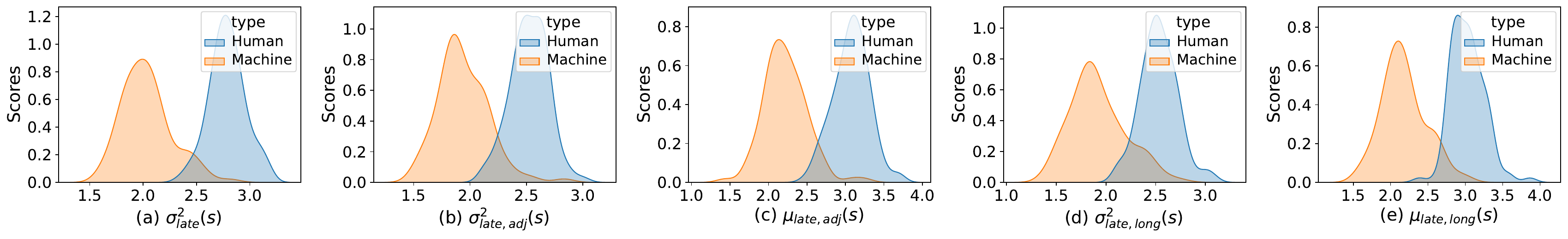}
 \vspace{-0.4cm}
	\caption{Visualization of the global statistics for detection scores of MGT and HGT. Here, DetectLLM scores are used and $t_0=20$. (a) demonstrates Score Stability; (b)--(c) demonstrate Adjacent Difference Stability; and (d)--(e) demonstrate Long-Range Difference Stability.}
	\label{fig: motivation_position_DetectLLM_Essay}
\end{figure*}

The local analysis stated above focuses on the adjacent case, i.e., one-step contextual transitions. However, contextual dependence in MGT is not limited to immediate neighbors. Token-level scores may also exhibit structured relations over longer ranges. The following theorem extends the local result to multi-hop score transitions.

\begin{theorem}
Let $d(s_t)=\mathcal{G}(\alpha_t)$ denote the token-level detection score at the $t$-th step, where $\mathcal{G}$ is $L$-Lipschitz with respect to the $L_{\infty}$-norm. Let $\lambda_K, \lambda_Q, \lambda_V, \lambda_O$ be the largest singular values of parameters $W_K, W_Q, W_V, W_O$, respectively, and let $W=W_V W_O W_Q W_K^{\top}$. For the transformer defined in Eq. (\ref{eq: simplified_model}), assuming normalized inputs ($\left\|x_t\right\|_2=1$ for all $t$) and constants $c, \epsilon>0$, consider $a_t x_{t+1}^{\top} \geq(1-\delta)\left\|a_t\right\|_2$ with $\delta \leq\left(\frac{c \epsilon}{\lambda_Q \lambda_K \lambda_V \lambda_O}\right)^2$, where $a_t=\alpha_tX_{t-1}W_VW_O$. If $x_{\ell}$ satisfies $x_{\ell} W x_{\ell}^{\top} \geq c$ and $x_{\ell} W x_{\ell} \geq \epsilon^{-1} \max _{j \in[\ell], j \neq \ell} x_j W x_{\ell}^{\top}$, then
\begin{equation}
\left|d(s_{t+k})-d(s_t)\right| \leq \frac{L(1+\sqrt{2}) \epsilon \cdot \max _j\left(x_j W x_j^T\right)\cdot \ln \left(1+\frac{k}{t}\right)}{\min_{0\le i< k} \|a_{t+i}\|_2}
\label{eq: global_bound}
\end{equation}
\label{theorem: global_relation}
\end{theorem}

Theorem \ref{theorem: global_relation} extends Theorem \ref{theorem: local_relation} from one-step transitions to multi-hop score differences. Its key implication is that even long-range score changes remain structurally bounded rather than arbitrary. This motivates us to move beyond local smoothness and examine how token-level evidence is globally organized across the entire text.

Ignoring the unstable initial part, i.e., when $t$ is large, $\ln(1+k/t)$ in Eq. \ref{eq: global_bound} tends toward 0, implying the score transitions of MGT are expected to be more tightly constrained. Instead, HGT is not constrained by the same autoregressive generation trajectory of a fixed LLM, and therefore may exhibit more dispersed score variations. Guided by this observation, we focus on three global properties.

\begin{itemize}[leftmargin=*]
\item \textbf{Score Stability}, which is measured by the variance of token-level scores in the latter part:  $\sigma^2_{\mathrm{late}}(s)=\mathrm{Var}(\{d(s_t)\}_{t>t_0})$, where $t_0$ is the predefined initial part length. The theorem motivates the expectation that score differences in the latter part of MGT tend to be more constrained (since $ln(1+k/t)\to0$ as $t$ is large) and thus exhibit lower variance. Empirically, Fig. \ref{fig: motivation_position_DetectLLM_Essay} (a) shows that MGT indeed exhibits lower score variance.

\item \textbf{Adjacent Difference Stability}, which is measured by the variance and mean of adjacent score difference in the latter part: $\sigma^2_{\mathrm{late, adj}}(s)=\mathrm{Var}(\Delta_{t>t_0}^{(1)})$, and $\mu_{\mathrm{late, adj}}(s)=\mathrm{Mean}(\Delta_{t>t_0}^{(1)})$, where $\Delta_t^{(1)} = \left|d(s_{t+1})-d(s_t)\right|$. Since the local case ($k=1$) of Theorem \ref{theorem: global_relation} is also tighter at later positions ($t$ is large), this statistic in MGT is expected to exhibit a lower value. As with Score Stability, the diversity inherent in HGT results in higher values for these metrics. As shown in Fig. \ref{fig: motivation_position_DetectLLM_Essay} (b)--(c), MGT again exhibits lower values, indicating more stable evolution in the latter part of the sequence.

\item \textbf{Long-Range Difference Stability}, which is measured by the variance and mean of long-range score difference in the latter part: $\sigma^2_{\mathrm{late, long}}(s)=\mathrm{Var}(\Delta_{t>t_0}^{(|N-t_0|/2)})$, and $\mu_{\mathrm{late, long}}(s)=\mathrm{Mean}(\Delta_{t>t_0}^{(|N-t_0|/2)})$, where $\Delta_t^{(|N-t_0|/2)} = \left|d(s_{t+|N-t_0|/2})-d(s_t)\right|$. The theorem suggests that, even when the hop size (i.e., $k$) is large, the fact that $t$ is also large makes these long-range differences relatively small and stable. Fig. \ref{fig: motivation_position_DetectLLM_Essay} (d)--(e) confirms that MGT displays smaller and more stable long-range score differences than HGT, revealing a distinctive global organization of token-level score.
\end{itemize}

These findings complement the local results in Subsection \ref{sec: local}. Local contextual relations describe short-range smoothness and early-position instability, whereas global contextual relations reveal how token-level evidence is organized over the entire text.

\section{Multi-Level Contextual Relation Modeling}

\begin{figure*}[t]
	\centering
	\includegraphics[width=0.95\linewidth]{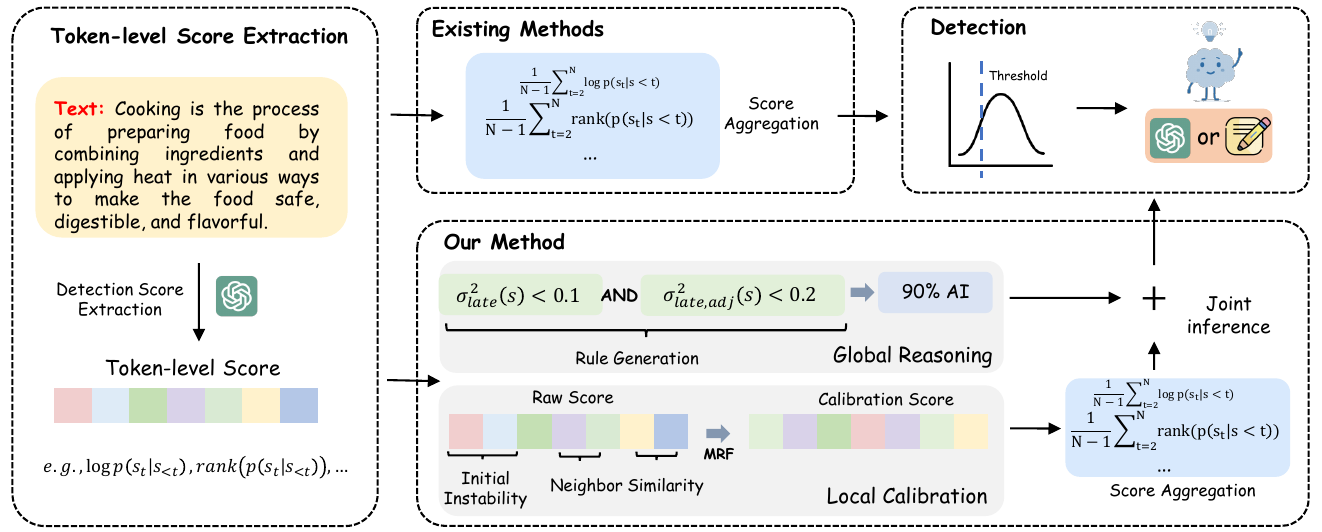}
 \vspace{-0.1cm}
	\caption{The workflow of the Multi-level contextual token relation modeling framework.}
	\label{fig: framework}
\end{figure*}

The previous section shows that token-level detection scores exhibit both local and global contextual relations. Motivated by these observations, we develop a multi-level framework with three components: a local calibration module for short-range score refinement, a global rule-support reasoning module for sequence-level pattern modeling, and a joint inference mechanism that combines the two signals. The framework is shown in Fig. \ref{fig: framework}.

\subsection{Local Relation Modeling via Markov Calibration}
\label{sec: local_modeling}

Section \ref{sec: local} reveals two local properties of token-level detection scores: Neighbor Similarity and Initial Instability. To capture these properties, we adopt a lightweight Markov calibration module that refines token scores before token-level score aggregation.

\subsubsection{Markov Random Field for MGT Detection}
For each token $s_t$ in text $s$, we assign a binary random variable $y_{s_{t}}$, where $y_{s_{t}}=0$ and $y_{s_{t}}=1$ indicate a human- or machine-generated token\footnote{Note that token labels are not absolute but depend on the context in which they appear. For example, "the" can be a human token or a machine token depending on the text.}, respectively, as measured by the detection score of the token. Let $y_{s}$ denote the label set for all tokens in text $s$, the pMRF over these tokens can be formalized as a Gibbs distribution \cite{gao2022bayesian}:
$P(y_s)=\frac{1}{Z} \exp(-E(s,y_s)),$
where $Z$ is a normalizing constant and $E(s, y_s)$ is the energy function. Our objective is to maximize the posterior probability of the token labels $y_s$ by minimizing the global energy function $E(s,y_s)$. The energy function typically consists of two components: the unary potential $\Psi_U$ and the pairwise potential $\Psi_P$:
$$E(s,y_s)=\sum\limits_{t=1}^{N} \Psi_U\left(s_t, y_{s_t}\right)+\sum\limits_{t=1}^{N}\sum\limits_{s_j\in\mathcal{N}(s_t)} \Psi_P\left(y_{s_t}, y_{s_{j}}\right),$$
where $\mathcal{N}(s_t)=\{s_{t-1},s_{t+1}\}$ denotes the adjacent tokens of token $s_t$.

\textbf{Unary potential $\Psi_U\left(s_t, y_{s_t}\right)$} quantifies the cost of assigning label $y_{s_t}$ to token $s_t$. We let $\Psi_U\left(s_t, y_{s_t}\right)=-\log p(s_t)$, where $p(s_t)$ is the output probability from the original detector, which is measured by the 0-1 normalized detection score of token $s_t$, i.e., the normalized $d(s_t)$.

\textbf{Pairwise potential $\Psi_P\left(y_{s_t}, y_{s_{j}}\right)$} models the similarity in detection scores between adjacent tokens. A penalty is applied if two adjacent tokens are assigned diﬀerent labels; otherwise, a reward is given. This enforces label smoothness and captures the neighbor similarity property:
\begin{equation}
\label{eq: pairwise_potential_temp}
    \Psi_P\left(y_{s_t}, y_{s_{j}}\right)=w\cdot(2\cdot I(y_{s_t}\ne y_{s_{j}})-1),
\end{equation}
where $I(\cdot)$ is the indicator, and the reward and penalty factor $w\ge 0$. This implies an energy penalty of $w$ when adjacent tokens have different labels; otherwise, the reward is $-w$.

To model the initial instability property, we introduce a positional weighting function $\beta(t)$ in the binary potential. This function assigns lower weights to binary potentials at earlier positions, thereby mitigating the errors caused by unstable initial neighbor tokens. In this paper, we define the positional weighting function $\beta(t)$ as a Sigmoid function to ensure a smooth transition of weights, and the revised binary potential is then given by:
\begin{equation}
\begin{gathered}
\Psi_P\left(y_{s_i}, y_{s_j}\right)=\beta(j) \cdot w \cdot\left(2 \cdot I\left(y_{s_i} \neq y_{s_j}\right)-1\right), \\
\text { with } \beta(j)=\frac{1}{1+\exp \left(-\left(j-t_0\right)\right)},
\end{gathered}
\label{eq: pairwise_potential}
\end{equation}
where $t_0$ is the predefined initial part length, effectively suppressing the pairwise potential of tokens before $t_0$.

\subsubsection{Mean Field Approximation in MGT Detection}
Exact inference in this MRF is intractable. We therefore adopt a mean-field approximation with a fully factorized variational distribution $Q(y)=\prod_{t=1}^{N}Q_t(y_t)$. After standard derivation (details can be found in Section II of Supplementary Material), the resulting update can be written in matrix form as

\begin{equation}
    Q^{(r)}=
\mathrm{softmax}\!\left(
\log Q^{(r-1)}
-
AQ^{(r-1)}
\Big(
W\odot
\begin{bmatrix}
-1 & 1\\
1 & -1
\end{bmatrix}
\Big)
\right),
\label{eq: iter_update}
\end{equation}
with $Q^{(0)}=[1-p(s),p(s)]$, where $p(s)=[p(s_1),\ldots,p(s_N)]^ \mathrm{ T }$. For the adjacent matrix $A$, $A_{t,t+1}=\beta(t+1)$ for all $t=0,...,N-2$, $A_{t-1,t}=\beta(t-1)$ for all $t=1,...,N-1$, and 0 otherwise. $W\in\mathbb{R}_{+}^{2\times2}$ is the reward and punishment weights. This iterative refinement propagates reliable local evidence across neighboring tokens while suppressing unstable early interactions.

After $T$ iterations, we further downweight early positions in the final calibrated scores:
\begin{equation}
    Q^{\mathrm{final}}=
\mathrm{Diag}\!\big(\beta(1),\dots,\beta(N)\big)\,Q^{(T)}.
\label{eq: final_calibration}
\end{equation}
The calibrated token scores are then passed to the original detector's aggregation component. If the base detector is decomposed into a token-level scoring module $f_{\mathrm{tok}}$ and a text-level decision module $f_{\mathrm{dec}}$, the enhanced detector becomes
\[
f_{\mathrm{enh}}(s)=f_{\mathrm{dec}}\!\big(f_{\mathrm{mrf}}(f_{\mathrm{tok}}(s))\big),
\]
where $f_{\mathrm{mrf}}$ denotes the Markov calibration module of Eq. \ref{eq: final_calibration}.

\subsection{Global Relation Modeling via Rule-support Reasoning}
\label{sec: global_model}

Subsection \ref{sec: global} shows that MGT exhibits global relations, including lower score variance, lower adjacent-difference variance, and lower long-range difference variance. These global patterns are difficult to capture by local smoothing alone. Given the inferential power of symbolic logic \cite{duan2022deeplogic}, we introduce a rule-support reasoning module.

For a detector estimating the probability $p(y\mid s)$, we introduce a latent variable $\alpha$ to denote the logical rule. This yields the following ideal rule-reasoning formulation:
\begin{equation}
\begin{aligned}
    p(y \mid s, b)&=\sum\nolimits_{\alpha} p(y \mid \alpha, s, b) p(\alpha \mid s, b),
\end{aligned}
\label{eq: rule_reason}
\end{equation}
where $b$ represents the prior knowledge about the rules, e.g., the desirable form of rules. We can further decompose it as follows (proof in Section III of Supplementary Material):
\begin{equation}
\begin{aligned}
p(y \mid s, b)\propto \sum_{\alpha} \underbrace{p(b \mid \alpha)}_{\substack{\text { Rule Prior }}} \cdot \underbrace{p(y \mid \alpha)}_{\substack{\text { Detection }}} \cdot \underbrace{p(\alpha \mid s)}_{\substack{\text {Rule Generation}}}.
\end{aligned}
\end{equation}

The three derived terms correspond to three main parts of the rule-support reasoning module.

\subsubsection{Rule Prior $p(b\mid\alpha)$}
The rule priors are used to constrain the rule employed for feasibility detection. In the paper, they are treated as global statistics identified in the previous section, e.g., $\sigma^2_{late}(s)$ and $\sigma^2_{late, adj}(s)$. More generally, suppose each text $s$ is represented by $M$ global statistics:
\[
z(s)=\big[z_1(s),z_2(s),\dots,z_M(s)\big].
\]
For each statistic $z_m(s)\in z(s)$, we compute its empirical range on the training set $\mathcal{D}_{train}$ and uniformly divide it into $K$ intervals using $K-1$ thresholds $\{\tau_{m,j}\}_{j=1}^{K-1}$:
\[
\tau_{m,j}=a_m+\frac{j-1}{K-1}(b_m-a_m), \qquad j=1,\dots,K-1,
\]
where
\[
a_m=\min_{s\in \mathcal{D}_{\mathrm{train}}} z_m(s), \qquad
b_m=\max_{s\in \mathcal{D}_{\mathrm{train}}} z_m(s).
\]

Then, for each statistic $z_m(s)$, we obtain K threshold atoms (the smallest unit of the rule):
\[
z_m(s)\le \tau_{m,1},\;
\tau_{m,1} \le z_m(s)\le \tau_{m,2},\;
\dots,\;
z_m(s) > \tau_{m,K-1},
\]
and similarly for the other statistics.
Based on these threshold atoms, we define the feasible rule space $\Omega$ using "AND" conjunction, i.e., a rule takes the form "atom 1 AND atom 2 AND ...". 

Naturally, a candidate rule $\alpha$ is feasible (i.e. $\alpha\in\Omega(\alpha)$) only if it satisfies the following constraints:
(1) each atom in $\alpha$ must be one of the threshold atoms defined above;
(2) exactly one threshold atom is selected from each statistic, so as to avoid impossible rules such as
$z_m(s)\le \tau_{m,1}$ AND $\tau_{m,1}\le z_m(s)\le \tau_{m,2}$;
(3) the number of threshold atoms in the rule $\alpha$ is $M$.
Therefore, the rule prior is
\begin{equation}
    p_h(b\mid \alpha)=
\begin{cases}
1, & \alpha \in \Omega(\alpha),\\
0, & \text{otherwise}.
\end{cases}
\label{eq: prior}
\end{equation}

\subsubsection{Rule Generation $p(\alpha\mid s)$}

For a given input text $s$, each statistic $z_m(s)$ falls into one interval and therefore activates exactly one threshold atom, denoted by $o_m(s)$. We then generate the rule for $s$ deterministically as
\[
\alpha(s)=o_1(s)\ \text{AND}\ o_2(s) \text{AND}\ ...\ \text{AND}\  o_m(s).
\]
Thus, rule generation is a direct mapping from the global statistic to a deterministic rule, and correspondingly, the rule-generation distribution degenerates to a one-hot form centered at $\alpha(s)$, that is,
\begin{equation}
    p(\alpha \mid s)=
\begin{cases}
1, & \alpha=\alpha(s),\\
0, & \text{otherwise}.
\end{cases}
\label{eq: rule_generation}
\end{equation}

\subsubsection{Detection $p(y\mid\alpha)$}

Ideally, the detection term $p(y\mid \alpha)$ can be estimated by obtaining all samples that satisfy rule $\alpha$ and then calculating the percentage of them that have label $y$, i.e.,
\[
\hat p(y\mid \alpha)=\frac{n_{\alpha,y}}{n_\alpha},
\]
where $n_\alpha$ is the number of training texts satisfying rule $\alpha$, and $n_{\alpha,y}$ is the number of such texts with class label $y$. 

However, this ideal estimation is unreliable in our setting. Since the training set contains only a few texts, while the number of possible conjunction rules grows combinatorially after discretization ($K^M$), many specific rules are supported by very few samples. As a result, the exact estimation of $p(y\mid\alpha)$ becomes highly sparse and unstable.

Therefore, in the actual implementation, we approximate the decision term by a stable rule-support score derived from the activated atomic rules. For an atom $o$, we compute its detection-support score as
\[
r(y\mid o)=
\begin{cases}
\frac{n_{o,y}}{n_o}, & n_o>0,\\[4pt]
\frac{1}{2}, & n_o=0,
\end{cases}
\]
where $n_o$ is the number of training texts satisfying atom $o$, $n_{o,y}$ is the number of such texts with label $y$. For unseen atoms, we use a neutral score.

Given the generated rule $\alpha(s)=\bigwedge_{m=1}^{M}o_m(s)$, we define the practical approximation of the detection term by aggregating the support scores of its activated atoms:
\[
r(y\mid \alpha(s))
=
\frac{1}{M}\sum_{m=1}^{M} r(y\mid o_m(s)).
\]

For the rule $\alpha(s)$ extracted from $s$, it is clear that $p(b\mid\alpha)=1$ and $p(\alpha\mid s)=1$, then the rule-support reasoning formulation is approximated in practice as:
\[
\tilde p(y\mid s,b)\propto r(y\mid \alpha(s)).
\]
In the binary setting of MGT detection, we use
\begin{equation}
    r_{\mathrm{rule}}(s)=r(MGT\mid \alpha(s))
\end{equation}
as the output of the rule-support reasoning module.

Notably, the rule score is not intended to be a calibrated posterior probability. Instead, it is a support score induced by global statistics. This design avoids the sparsity of full conjunction rules while preserving the semantics  of rule activation. We also compare with strict probabilistic rules in Section \ref{sec: ablation_probability} to demonstrate the rationality of our design.

\subsection{Joint Multi-Level Inference}
\label{sec: joing_inference}

The local and global modules play complementary roles. The local branch refines token-level evidence by exploiting local contextual consistency, while the global branch captures global contextual relations to provide a confident score.

Since $r_{\mathrm{rule}}(s)$ is a rule-support score rather than a strict probability, it is not used as a standalone prediction. Therefore, we treat it as a complementary confidence that adjusts the final detector score, rather than as a standalone prediction score. The final detection score is as follows
\begin{equation}
    F(s)=f_{\mathrm{enh}}(s)+\lambda \, r_{\mathrm{rule}}(s),
\end{equation}
where $\lambda \ge 0$ is a learnable coefficient that controls the contribution of the global rule confidence.

\textbf{Complexity Analysis}. The parameters of our method are learned from text-level supervision in the training set. For the local calibration module, since the update only involves sparse-dense matrix multiplications over adjacent positions, its computational complexity is $\mathcal{O}(NT)$ for a text of length $N$ and $T$ refinement iterations. For the global rule-support reasoning module, the complexity of computing global statistics is $\mathcal{O}(NM)$, determining the threshold atom to which each statistic belongs is $\mathcal{O}(MK)$, and aggregating the rule support score is $\mathcal{O}(M)$. Overall, the complexity is $\mathcal{O}(NT+NM+MK)$.

\textbf{Overall Framework}. Alg. \ref{alg:joint_inference} summarizes the overall inference procedure of the proposed framework. Specifically, the \textsc{Local Branch} (Lines 5-13) first computes raw token-level scores from the base detector, and then refines them through the Markov-informed calibration module. This produces the locally enhanced detection score $f_{\mathrm{enh}}(s)$. The \textsc{Global Branch} (Lines 14-23) then computes the global statistics of the input text, maps each statistic to its corresponding threshold atom, constructs a deterministic logical rule, and aggregates the activated atom supports into the rule score $r_{\mathrm{rule}}(s)$. Finally, the two branches are fused through a simple additive form to obtain the final score $F(s)$ (Line 3).

\begin{algorithm}[tb]
\caption{Joint Multi-Level Inference for MGT Detection}
\label{alg:joint_inference}
\begin{algorithmic}[1]
\Require Text $s=\{s_t\}_{t=1}^{N}$; base detector $f_{\mathrm{tok}},f_{\mathrm{dec}}$.
\Require Local calibration parameters $W_{\mathrm{mrf}}, t_0, T$; 
\Require Global statistic functions $\{z_m(\cdot)\}_{m=1}^{M}$; thresholds $\{\tau_{m,j}\}_{m=1,j=1}^{M,K-1}$; 
atom support scores $r(y\mid o)$; 
fusion weight $\lambda$.

\State $f_{\mathrm{enh}}(s)\gets \Call{Local\_Calibration}{s}$.
\State $r_{\mathrm{rule}}(s)\gets \Call{Global\_Reasoning}{s}$.
\State $F(s)\gets f_{\mathrm{enh}}(s)+\lambda\,r_{\mathrm{rule}}(s)$.
\State \Return detection score $F(s)$ of text $s$.

\Function{Local\_Calibration}{$s$}
    \State Compute raw token-level scores $p(s)=f_{\mathrm{tok}}(s)$.
    \State set $Q^{(0)}=[1-p(s),p(s)]$.
   \For{$r=1$ {\bfseries to} $T$}
   \State Update $Q^{(r)}$ according to Eq. \ref{eq: iter_update}.
   \EndFor
   \State Calculate $Q_{\text{final}}$ according to Eq. \ref{eq: final_calibration}.
   \State \Return calibrated score $f_{\mathrm{enh}}(s)=f_{\mathrm{dec}}(Q^{\mathrm{final}})$.
\EndFunction

\Function{Global\_Reasoning}{$s$}
    \State $z(s)\gets [z_1(s),z_2(s),\dots,z_M(s)]$
    \For{$m=1$ to $M$}
        \State Determine the activated threshold atom of $z_m(s)$ according to $\{\tau_{m,j}\}_{j=1}^{K-1}$.
        \State Activate the corresponding atom $o_m(s)$.
    \EndFor
    \State $\alpha(s)\gets \bigwedge_{m=1}^{M} o_m(s)$.
    \State $r_{\mathrm{rule}}(s)\gets \frac{1}{M}\sum_{m=1}^{M} r(\mathrm{MGT}\mid o_m(s))$.
    \State \Return $r_{\mathrm{rule}}(s)$.
\EndFunction

\end{algorithmic}
\end{algorithm}

\section{Experiments}

\subsection{Experimental Settings}
\subsubsection{Datasets}
We evaluate the proposed method on four public datasets:
\begin{itemize}[leftmargin=*]
    \item \textbf{Essay} \cite{verma2024ghostbuster}. Each source contains 1,000 samples. The HGT portion comprises IvyPanda essays across various academic levels. The corresponding MGTs were generated by prompting multiple LLMs (including GPT4All, ChatGPT, ChatGPT‑turbo, ChatGLM, Dolly, and Claude) based on topics from the source documents.
    \item \textbf{Reuters} \cite{verma2024ghostbuster}. Built on the Reuters 50–50 authorship benchmark, this dataset contains 1,000 articles from 50 writers. ChatGPT‑turbo was used to invent a headline for every article. Those auto-generated headlines were then embedded into prompts and submitted to multiple LLMs, including ChatGPT, GPT‑4, ChatGPT‑turbo, ChatGLM, Dolly, and Claude, to create the machine-generated texts.
    \item \textbf{TruthfulQA} \cite{lin2022truthfulqa}. It contains 817 questions covering 38 categories, including health, law, finance, and politics. The generated answers were produced by several large language models, including GPT4, ChatGPT-turbo, ChatGLM, Dolly, ChatGPT, and StableLM.
    \item \textbf{DetectRL} \cite{wu2024detectrl}. HGTs include 2,800 samples each from arXiv, XSum, Writing Prompts, and Yelp. MGTs were generated by ChatGPT, PaLM-2, and Llama-2. It also features practical adversarial settings, including paraphrasing attacks \cite{krishna2023paraphrasing} and mixed-text conditions. The dataset further models practical adversarial settings: (1) paraphrasing attacks that rewrites MGTs with the Dipper \cite{krishna2023paraphrasing} and Polish paraphraser, and (2) a mixed-text condition where 1/4 of machine-generated sentences is randomly replaced with human-written content while the label remains MGT.
\end{itemize}

\subsubsection{Baselines}
We choose the following detection methods for base detectors for enhancements.
\begin{itemize}[leftmargin=*]
    \item \textbf{Likelihood} \cite{solaiman2019release}. It uses an LLM to calculate the log probability of each token in a text. The average of these probabilities gives a detection score. A higher score indicates a greater likelihood that the text was generated by an LLM.
    \item {\textbf{Log-Rank} \cite{gehrmann2019gltr}}. Its detection score is computed by first using an LLM to rank each token in a text according to its predicted order in the given context. The logarithm of each word's predicted rank is then calculated. The final score is an average of these values, and a lower score is a strong indicator of machine-generated text.
    \item \textbf{Entropy} \cite{gehrmann2019gltr}. Similar to Log-Rank, it calculates a score for a text by taking the average of each token's conditional entropy within its given context. A lower score indicates a higher likelihood that the text was generated by an LLM.
    \item \textbf{DetectGPT} \cite{mitchell2023detectgpt}. Its underlying idea is that text created by LLMs is already a high-probability output. So, when it is slightly altered, the new version is likely to have a lower log probability. In contrast, making similar small changes to human-written text does not consistently lower the log probability; it can just as easily stay the same or increase.
    \item \textbf{Fast-DetectGPT (FastGPT)} \cite{bao2024fast}. To overcome the major computational expense of DetectGPT, this approach replaces DetectGPT's resource-intensive perturbation step with a more efficient sampling process. It identifies differences in token selection between humans and LLMs using a conditional probability curvature metric.
    \item \textbf{Binoculars} \cite{hans2024spotting}. It is a detection algorithm that requires no training data and accurately distinguishes between human- and machine-generated text by comparing the score differences between a pair of pre-trained LLMs.
    \item \textbf{FourierGPT} \cite{xu2024detecting}. It proposes a detection method from a likelihood spectrum perspective, capturing subtle differences between MGT and HGT by analyzing relative changes in text likelihood values rather than their absolute values.
    \item \textbf{AdaDetectGPT (AdaGPT) \cite{zhouadadetectgpt}}. To overcome the suboptimality of existing logic-based detectors, which rely solely on log-probabilities, this method introduces a classifier that adaptively learns witness functions from training data.
    \item \textbf{DNA-DetectLLM (DetectLLM) \cite{zhudna}}. Based on a DNA-inspired perspective, it detects MGT through a repair-based process. It constructs an ideal AI-generated sequence, iteratively repairs non-optimal tokens, and quantifies the cumulative repair effort as an interpretable detection signal.
\end{itemize}

The versions utilizing the preliminary work and the proposed method are designated by the suffixes 'M' and 'Mult', respectively, e.g., DetectGPT-M and DetectGPT-Mult.

\begin{table*}[htbp]
  \centering   \caption{Performance concerning TPR@FPR-1\% (\%) on Essay (left) and TruthfulQA (right). The detectors are trained on GPT4All texts on Essay and GPT4 texts on TruthfulQA.}   \renewcommand\arraystretch{1.1}   \begin{adjustbox}{width=0.9\textwidth}   \setlength{\tabcolsep}{0.008\linewidth}
    \begin{tabular}{c|ccccc|c|ccccc|c}
    \toprule
    \multirow{2}[0]{*}{Detector} & \multicolumn{6}{c|}{Essay}                     & \multicolumn{6}{c}{TruthfulQA} \\
    \noalign{\smallskip} \cline{2-13}\noalign{\smallskip} 
          & GPT4All  & ChatGPT  & ChatGLM  & Dolly  & Claude  & Avg   & ChatGLM  & Dolly  & ChatGPT  & GPT4  & StableLM  & Avg \\
    \midrule
    Likelihood & $46.33_{\pm16.49}$ & $68.62_{\pm13.32}$ & $92.86_{\pm4.84}$ & $20.67_{\pm10.79}$ & $12.36_{\pm6.32}$ & 52.41 & $73.54_{\pm7.83}$ & $17.74_{\pm2.72}$ & $52.01_{\pm10.78}$ & $38.45_{\pm3.34}$ & $50.14_{\pm8.09}$ & 46.38 \\
    Likelihood-M & $87.47_{\pm3.42}$ & $90.36_{\pm2.93}$ & $97.19_{\pm1.11}$ & $55.47_{\pm5.03}$ & $\textbf{40.44}_{\pm7.83}$ & 77.87 & $75.91_{\pm17.62}$ & $52.17_{\pm2.62}$ & $42.04_{\pm19.77}$ & $29.17_{\pm11.92}$ & $54.79_{\pm10.29}$ & 50.82 \\
    \rowcolor[rgb]{ .949,  .949,  .949} \textbf{Likelihood-Mult} & $\textbf{98.00}_{\pm0.86}$ & $\textbf{95.47}_{\pm2.23}$ & $\textbf{98.93}_{\pm0.70}$ & $\textbf{82.24}_{\pm6.10}$ & $34.98_{\pm5.62}$ & \textbf{82.65} & $\textbf{90.61}_{\pm2.34}$ & $\textbf{57.47}_{\pm5.69}$ & $\textbf{77.62}_{\pm9.82}$ & $\textbf{45.10}_{\pm5.00}$ & $\textbf{75.69}_{\pm5.94}$ & \textbf{69.30} \\
    \midrule
    Log-Rank & $62.69_{\pm13.36}$ & $79.07_{\pm7.89}$ & $95.71_{\pm2.05}$ & $25.11_{\pm8.35}$ & $19.20_{\pm8.26}$ & 60.44 & $83.22_{\pm4.08}$ & $16.45_{\pm5.07}$ & $54.50_{\pm2.09}$ & $40.17_{\pm4.46}$ & $59.26_{\pm2.20}$ & 50.72 \\
    Log-Rank-M & $87.38_{\pm2.50}$ & $88.89_{\pm2.84}$ & $98.04_{\pm0.43}$ & $52.12_{\pm4.18}$ & $\textbf{31.69}_{\pm4.15}$ & 75.39 & $\textbf{90.61}_{\pm2.11}$ & $44.48_{\pm4.99}$ & $83.57_{\pm1.73}$ & $\textbf{47.74}_{\pm1.72}$ & $79.26_{\pm2.52}$ & 69.13 \\
    \rowcolor[rgb]{ .949,  .949,  .949} \textbf{Log-Rank-Mult} & $\textbf{96.49}_{\pm0.88}$ & $\textbf{93.29}_{\pm2.34}$ & $\textbf{98.44}_{\pm0.65}$ & $\textbf{75.89}_{\pm10.97}$ & $30.62_{\pm6.82}$ & \textbf{79.51} & $89.46_{\pm8.36}$ & $\textbf{53.65}_{\pm8.84}$ & $\textbf{84.53}_{\pm7.34}$ & $46.48_{\pm4.55}$ & $\textbf{81.47}_{\pm6.93}$ & \textbf{71.12} \\
     \midrule
    Entropy & $2.73_{\pm0.69}$ & $7.16_{\pm3.17}$ & $6.88_{\pm3.05}$ & $2.29_{\pm1.36}$ & $3.91_{\pm2.24}$ & 6.04  & $60.48_{\pm15.23}$ & $18.38_{\pm6.58}$ & $38.02_{\pm9.81}$ & $24.30_{\pm4.47}$ & $32.29_{\pm14.76}$ & 34.69 \\
    Entropy-M & $15.63_{\pm0.87}$ & $41.24_{\pm2.25}$ & $40.85_{\pm4.37}$ & $16.95_{\pm3.12}$ & $\textbf{23.20}_{\pm1.70}$ & 31.3  & $84.65_{\pm18.12}$ & $52.70_{\pm13.35}$ & $78.87_{\pm9.18}$ & $39.94_{\pm3.25}$ & $69.52_{\pm11.82}$ & 65.13 \\
    \rowcolor[rgb]{ .949,  .949,  .949} \textbf{Entropy-Mult} & $\textbf{91.89}_{\pm13.75}$ & $\textbf{81.91}_{\pm31.21}$ & $\textbf{97.46}_{\pm3.67}$ & $\textbf{91.55}_{\pm13.68}$ & $9.78_{\pm4.64}$ & \textbf{64.60} & $\textbf{94.84}_{\pm1.35}$ & $\textbf{58.40}_{\pm6.00}$ & $\textbf{86.12}_{\pm5.09}$ & $\textbf{46.82}_{\pm4.17}$ & $\textbf{79.43}_{\pm4.92}$ & \textbf{73.12} \\
     \midrule
    DetectGPT & $0.00_{\pm0.00}$ & $0.00_{\pm0.00}$ & $0.00_{\pm0.00}$ & $0.00_{\pm0.00}$ & $0.89_{\pm0.24}$ & 0.15  & $60.82_{\pm6.22}$ & $13.75_{\pm3.45}$ & $52.41_{\pm7.71}$ & $21.66_{\pm7.94}$ & $39.49_{\pm4.15}$ & 37.63 \\
    DetectGPT-M & $41.23_{\pm29.53}$ & $46.93_{\pm25.36}$ & $41.83_{\pm44.80}$ & $18.38_{\pm20.21}$ & $8.80_{\pm9.64}$ & 37.17 & $0.29_{\pm0.44}$ & $1.06_{\pm1.35}$ & $0.34_{\pm0.55}$ & $1.03_{\pm0.86}$ & $0.06_{\pm0.11}$ & 0.55 \\
    \textbf{DetectGPT-Mult} & $\textbf{97.86}_{\pm0.81}$ & $\textbf{95.60}_{\pm2.17}$ & $\textbf{98.84}_{\pm0.99}$ & $\textbf{82.15}_{\pm6.16}$ & $\textbf{33.11}_{\pm4.56}$ & \textbf{82.32} & $\textbf{88.66}_{\pm5.43}$ & $\textbf{55.82}_{\pm5.50}$ & $\textbf{86.69}_{\pm6.60}$ & $\textbf{41.09}_{\pm9.49}$ & $\textbf{79.60}_{\pm7.33}$ & \textbf{70.37} \\
     \midrule
    FastGPT & $1.59_{\pm0.32}$ & $1.64_{\pm0.67}$ & $2.72_{\pm0.68}$ & $0.29_{\pm0.10}$ & $3.47_{\pm1.83}$ & 4.43  & $66.61_{\pm6.47}$ & $20.63_{\pm4.37}$ & $52.58_{\pm9.86}$ & $21.20_{\pm4.54}$ & $49.92_{\pm6.27}$ & 42.19 \\
    FastGPT-M & $23.92_{\pm10.29}$ & $30.00_{\pm9.20}$ & $55.54_{\pm7.80}$ & $32.89_{\pm16.52}$ & $0.67_{\pm0.56}$ & 26.58 & $83.27_{\pm8.05}$ & $30.31_{\pm11.96}$ & $73.94_{\pm8.22}$ & $35.76_{\pm2.49}$ & $63.46_{\pm11.25}$ & 57.35 \\
    \rowcolor[rgb]{ .949,  .949,  .949} \textbf{FastGPT-Mult} & $\textbf{98.72}_{\pm1.09}$ & $\textbf{97.91}_{\pm1.33}$ & $\textbf{99.60}_{\pm0.30}$ & $\textbf{92.74}_{\pm5.99}$ & $\textbf{40.27}_{\pm7.96}$ & \textbf{85.25} & $\textbf{89.46}_{\pm3.99}$ & $\textbf{45.47}_{\pm8.98}$ & $\textbf{83.97}_{\pm3.64}$ & $\textbf{43.15}_{\pm2.79}$ & $\textbf{74.62}_{\pm4.28}$ & \textbf{67.33} \\
     \midrule
    Binoculars & $94.94_{\pm0.81}$ & $\textbf{97.47}_{\pm0.76}$ & $\textbf{99.60}_{\pm0.17}$ & $73.89_{\pm3.91}$ & $44.62_{\pm8.20}$ & 84.74 & $8.08_{\pm10.78}$ & $1.88_{\pm0.40}$ & $3.68_{\pm2.50}$ & $6.02_{\pm4.96}$ & $6.86_{\pm5.15}$ & 5.3 \\
    Binoculars-M & $97.72_{\pm0.38}$ & $97.38_{\pm0.36}$ & $99.38_{\pm0.26}$ & $80.24_{\pm2.78}$ & $\textbf{59.51}_{\pm1.42}$ & \textbf{88.95} & $46.65_{\pm23.33}$ & $12.86_{\pm7.23}$ & $10.20_{\pm13.68}$ & $17.59_{\pm10.97}$ & $24.08_{\pm15.06}$ & 22.28 \\
    \rowcolor[rgb]{ .949,  .949,  .949} \textbf{Binoculars-Mult} & $\textbf{98.09}_{\pm0.78}$ & $96.62_{\pm1.68}$ & $99.11_{\pm0.53}$ & $\textbf{83.77}_{\pm5.66}$ & $38.13_{\pm5.07}$ & 84.47 & $\textbf{87.80}_{\pm6.10}$ & $\textbf{53.89}_{\pm8.08}$ & $\textbf{70.99}_{\pm11.65}$ & $\textbf{44.70}_{\pm7.10}$ & $\textbf{74.79}_{\pm7.12}$ & \textbf{66.43} \\
     \midrule
    FourierGPT & $98.13_{\pm1.65}$ & $91.60_{\pm4.43}$ & $98.35_{\pm1.95}$ & $97.61_{\pm3.58}$ & $0.00_{\pm0.00}$ & 64.29 & $96.28_{\pm0.87}$ & $69.51_{\pm2.02}$ & $92.24_{\pm1.72}$ & $56.33_{\pm1.29}$ & $89.12_{\pm1.39}$ & 80.7 \\
    FourierGPT-M & $95.03_{\pm9.82}$ & $96.80_{\pm3.99}$ & $99.69_{\pm0.30}$ & $98.52_{\pm1.95}$ & $0.00_{\pm0.00}$ & 65.01 & $96.45_{\pm0.62}$ & $\textbf{70.10}_{\pm1.44}$ & $92.69_{\pm1.26}$ & $56.28_{\pm1.10}$ & $89.18_{\pm1.26}$ & 80.94 \\
    \rowcolor[rgb]{ .949,  .949,  .949} \textbf{FourierGPT-Mult} & $\textbf{99.91}_{\pm0.18}$ & $\textbf{99.64}_{\pm0.39}$ & $\textbf{99.96}_{\pm0.09}$ & $\textbf{98.62}_{\pm1.41}$ & $\textbf{1.82}_{\pm1.14}$ & \textbf{67.47} & $\textbf{96.56}_{\pm0.83}$ & $69.51_{\pm0.93}$ & $\textbf{93.14}_{\pm1.07}$ & $\textbf{56.79}_{\pm1.79}$ & $\textbf{89.52}_{\pm1.43}$ & \textbf{81.10} \\
     \midrule
    AdaGPT & $22.23_{\pm3.90}$ & $20.67_{\pm4.68}$ & $46.52_{\pm7.22}$ & $30.98_{\pm4.27}$ & $0.76_{\pm0.59}$ & 20.3  & $71.99_{\pm10.88}$ & $36.02_{\pm3.36}$ & $47.31_{\pm7.46}$ & $19.54_{\pm1.73}$ & $50.76_{\pm4.38}$ & 45.12 \\
    AdaGPT-M & $26.15_{\pm2.28}$ & $26.09_{\pm6.81}$ & $52.54_{\pm6.12}$ & $38.81_{\pm3.84}$ & $0.71_{\pm0.47}$ & 24.17 & $88.72_{\pm1.50}$ & $42.95_{\pm2.11}$ & $64.76_{\pm7.13}$ & $30.20_{\pm2.70}$ & $70.54_{\pm3.06}$ & 59.43 \\
    \rowcolor[rgb]{ .949,  .949,  .949} \textbf{AdaGPT-Mult} & $\textbf{99.09}_{\pm0.66}$ & $\textbf{97.87}_{\pm1.35}$ & $\textbf{99.64}_{\pm0.23}$ & $\textbf{93.46}_{\pm4.33}$ & $\textbf{36.53}_{\pm4.59}$ & \textbf{84.11} & $\textbf{92.78}_{\pm2.17}$ & $\textbf{54.24}_{\pm8.24}$ & $\textbf{79.21}_{\pm12.78}$ & $\textbf{40.74}_{\pm3.45}$ & $\textbf{76.54}_{\pm10.01}$ & \textbf{68.70} \\
     \midrule
    DetectLLM & $3.01_{\pm0.57}$ & $1.02_{\pm0.68}$ & $1.25_{\pm0.46}$ & $1.96_{\pm0.66}$ & $0.09_{\pm0.18}$ & 1.27  & $1.20_{\pm0.75}$ & $1.70_{\pm0.73}$ & $0.57_{\pm0.25}$ & $0.86_{\pm0.51}$ & $1.08_{\pm0.42}$ & 1.08 \\
    DetectLLM-M & $15.44_{\pm16.68}$ & $17.96_{\pm27.41}$ & $25.49_{\pm33.42}$ & $12.41_{\pm15.32}$ & $5.47_{\pm10.71}$ & 15.29 & $0.51_{\pm0.52}$ & $1.65_{\pm0.71}$ & $0.28_{\pm0.31}$ & $0.40_{\pm0.23}$ & $0.62_{\pm0.49}$ & 0.69 \\
    \rowcolor[rgb]{ .949,  .949,  .949} \textbf{DetectLLM-Mult} & $\textbf{98.04}_{\pm0.74}$ & $\textbf{95.64}_{\pm2.72}$ & $\textbf{98.97}_{\pm0.76}$ & $\textbf{82.15}_{\pm6.56}$ & $\textbf{33.91}_{\pm3.89}$ & \textbf{82.32} & $\textbf{75.31}_{\pm17.23}$ & $\textbf{40.68}_{\pm17.65}$ & $\textbf{63.68}_{\pm31.02}$ & $\textbf{40.52}_{\pm9.12}$ & $\textbf{58.81}_{\pm26.48}$ & \textbf{55.80} \\
    \bottomrule
    \end{tabular}%
    \end{adjustbox}
  \label{tab: main_tpr}%
\end{table*}%

\subsubsection{Evaluation Metrics}
We evaluate detection performance using two metrics. First, we report the area under the ROC curve (\textbf{AUROC}) as the main metric for binary classification. Second, following recent MGT detection literature, we additionally report the true positive rate at a low false positive rate. It is particularly important in practice because falsely classifying human-written text as machine-generated can be highly undesirable. Specifically, we measure the TPR at an FPR of 1\%, denoting this as \textbf{TPR@FPR-1\%}.

\subsubsection{Experimental Protocol and Proxy Models}
We follow a strict black-box threat model for the source LLMs: the detector is assumed to be entirely unknown to the target generative model. Accordingly, we employ a proxy model to compute token-level scores for metric-based baselines. Here, GPT-2 is used as the proxy model for all baselines, while GPT-2-XL is used as the scoring model for Fast-DetectGPT and DetectLLM. For the trainable components of our method, we learn the local calibration and global rule-support reasoning parameters from the specified LLM-generated texts, then test on candidate texts generated by different LLMs, aiming to evaluate both in-LLM and cross-LLM generalization.

\subsubsection{Parameter Settings}
We conduct five independent runs using fixed random seeds \{1,2,3,4,5\}. For each dataset, 10\% of the data is used for training, and the remaining 90\% is evenly split into validation and test sets. The thresholds ${\tau_{m,j}}$ and atom-support scores $r(y\mid o)$ are estimated only on the training split. The fusion coefficient $\lambda$ and other tunable parameters are selected on the validation split, and the test split is used only for final evaluation. To ensure fair comparison, the enhanced detectors share the same hyperparameters as their corresponding base detectors. For the local calibration module, the number of MRF refinement iterations is set to $T=10$ by default. For the global rule-support reasoning module, the interval $K$ is set to 10. Besides, the initial part length $t_0$ is set to $t_0=20$. These values are kept fixed across datasets unless otherwise specified. 

\subsection{Performance Comparison}

In this section, we evaluate the effectiveness of the proposed method's enhancements in various real-world scenarios, including cross-LLM, cross-domain, mixed-text detection, and resistance to paraphrasing and adversarial attacks. The details of these scenarios are provided in Section IV of the Supplementary Material.

\begin{table*}[htbp]
  \centering   \caption{Performance concerning AUROC (\%) on Essay (left) and TruthfulQA (right). The detectors are trained on GPT4All texts on Essay and GPT4 texts on TruthfulQA.}   \renewcommand\arraystretch{0.9}   \begin{adjustbox}{width=1.\textwidth}   \setlength{\tabcolsep}{0.008\linewidth}
    \begin{tabular}{c|ccccc|c|ccccc|c}
    \toprule
    \multirow{2}[0]{*}{Detector} & \multicolumn{6}{c|}{Essay}                     & \multicolumn{6}{c}{TruthfulQA} \\
    \noalign{\smallskip} \cline{2-13}\noalign{\smallskip} 
          & GPT4All  & ChatGPT  & ChatGLM  & Dolly  & Claude  & Avg   & ChatGLM  & Dolly  & ChatGPT  & GPT4  & StableLM  & Avg \\
    \midrule
    Likelihood & $96.16_{\pm0.30}$ & $98.79_{\pm0.19}$ & $99.29_{\pm0.25}$ & $90.90_{\pm1.33}$ & $92.76_{\pm0.23}$ & 95.58 & $97.08_{\pm0.29}$ & $79.28_{\pm0.48}$ & $95.65_{\pm0.40}$ & $84.67_{\pm0.88}$ & $94.17_{\pm0.30}$ & 90.17 \\
    Likelihood-M & $98.58_{\pm0.14}$ & $99.47_{\pm0.11}$ & $99.54_{\pm0.18}$ & $94.59_{\pm0.96}$ & $\textbf{94.82}_{\pm0.36}$ & 97.4  & $97.14_{\pm0.92}$ & $83.70_{\pm0.57}$ & $94.20_{\pm1.87}$ & $86.61_{\pm0.56}$ & $94.37_{\pm0.92}$ & 91.2 \\
    \rowcolor[rgb]{ .949,  .949,  .949} \textbf{Likelihood-Mult} & $\textbf{99.92}_{\pm0.04}$ & $\textbf{99.81}_{\pm0.17}$ & $\textbf{99.95}_{\pm0.02}$ & $\textbf{99.18}_{\pm0.64}$ & $91.94_{\pm0.56}$ & \textbf{98.16} & $\textbf{97.72}_{\pm0.59}$ & $\textbf{84.94}_{\pm1.09}$ & $\textbf{96.98}_{\pm0.34}$ & $\textbf{87.29}_{\pm0.61}$ & $\textbf{96.20}_{\pm0.62}$ & \textbf{92.63} \\
    \midrule
    Log-Rank & $96.55_{\pm0.31}$ & $98.95_{\pm0.13}$ & $99.36_{\pm0.13}$ & $90.08_{\pm1.28}$ & $92.01_{\pm0.20}$ & 95.39 & $97.03_{\pm0.31}$ & $78.07_{\pm0.70}$ & $95.11_{\pm0.44}$ & $82.55_{\pm0.80}$ & $94.09_{\pm0.38}$ & 89.37 \\
    Log-Rank-M & $98.56_{\pm0.06}$ & $99.41_{\pm0.09}$ & $99.55_{\pm0.08}$ & $93.82_{\pm0.91}$ & $\textbf{92.91}_{\pm0.30}$ & 96.85 & $97.56_{\pm0.46}$ & $82.14_{\pm0.91}$ & $96.36_{\pm0.58}$ & $84.16_{\pm0.43}$ & $95.10_{\pm0.51}$ & 91.06 \\
    \rowcolor[rgb]{ .949,  .949,  .949} \textbf{Log-Rank-Mult} & $\textbf{99.86}_{\pm0.07}$ & $\textbf{99.71}_{\pm0.29}$ & $\textbf{99.93}_{\pm0.05}$ & $\textbf{98.79}_{\pm0.82}$ & $90.95_{\pm0.48}$ & \textbf{97.85} & $\textbf{97.74}_{\pm0.46}$ & $\textbf{84.18}_{\pm1.41}$ & $\textbf{96.88}_{\pm0.60}$ & $\textbf{84.92}_{\pm0.24}$ & $\textbf{95.72}_{\pm0.57}$ & \textbf{91.89} \\
    \midrule
    Entropy & $74.19_{\pm1.62}$ & $89.49_{\pm0.33}$ & $84.11_{\pm0.77}$ & $73.26_{\pm1.48}$ & $86.58_{\pm0.66}$ & 81.53 & $96.02_{\pm0.49}$ & $80.50_{\pm0.40}$ & $94.04_{\pm0.88}$ & $79.29_{\pm0.53}$ & $91.30_{\pm0.58}$ & 88.23 \\
    Entropy-M & $83.52_{\pm0.73}$ & $93.28_{\pm0.15}$ & $91.44_{\pm0.35}$ & $81.11_{\pm0.91}$ & $\textbf{87.96}_{\pm0.48}$ & 87.46 & $97.45_{\pm1.27}$ & $84.57_{\pm0.93}$ & $96.83_{\pm0.88}$ & $85.48_{\pm0.50}$ & $95.42_{\pm0.84}$ & 91.95 \\
    \rowcolor[rgb]{ .949,  .949,  .949} \textbf{Entropy-Mult} & $\textbf{99.64}_{\pm0.31}$ & $\textbf{96.98}_{\pm5.07}$ & $\textbf{99.80}_{\pm0.17}$ & $\textbf{98.62}_{\pm2.35}$ & $78.28_{\pm1.45}$ & \textbf{94.66} & $\textbf{97.96}_{\pm0.50}$ & $\textbf{85.84}_{\pm0.96}$ & $\textbf{97.50}_{\pm0.64}$ & $\textbf{86.65}_{\pm0.68}$ & $\textbf{96.32}_{\pm0.57}$ & \textbf{92.86} \\
    \midrule
    DetectGPT & $50.81_{\pm0.58}$ & $46.40_{\pm0.77}$ & $50.41_{\pm1.70}$ & $57.48_{\pm0.84}$ & $41.54_{\pm0.60}$ & 49.33 & $92.84_{\pm0.66}$ & $72.25_{\pm1.87}$ & $92.84_{\pm1.08}$ & $75.73_{\pm1.30}$ & $86.70_{\pm0.65}$ & 84.07 \\
    DetectGPT-M & $95.37_{\pm2.42}$ & $96.20_{\pm2.48}$ & $95.97_{\pm2.93}$ & $85.39_{\pm7.02}$ & $80.29_{\pm10.43}$ & 90.65 & $87.21_{\pm2.14}$ & $80.43_{\pm1.09}$ & $81.58_{\pm10.55}$ & $75.07_{\pm4.64}$ & $82.52_{\pm6.02}$ & 81.36 \\
    \rowcolor[rgb]{ .949,  .949,  .949} \textbf{DetectGPT-Mult} & $\textbf{99.93}_{\pm0.03}$ & $\textbf{99.78}_{\pm0.24}$ & $\textbf{99.95}_{\pm0.03}$ & $\textbf{98.97}_{\pm0.77}$ & $\textbf{90.17}_{\pm0.76}$ & \textbf{97.76} & $\textbf{98.23}_{\pm0.60}$ & $\textbf{87.48}_{\pm0.80}$ & $\textbf{97.49}_{\pm0.56}$ & $\textbf{85.32}_{\pm0.94}$ & $\textbf{96.43}_{\pm0.38}$ & \textbf{92.99} \\
    \midrule
    FastGPT & $64.63_{\pm1.53}$ & $67.68_{\pm1.70}$ & $71.08_{\pm1.51}$ & $47.17_{\pm1.53}$ & $75.31_{\pm0.90}$ & 65.17 & $96.26_{\pm0.29}$ & $81.63_{\pm0.76}$ & $94.38_{\pm0.65}$ & $82.61_{\pm1.26}$ & $91.84_{\pm0.57}$ & 89.34 \\
    FastGPT-M & $87.22_{\pm3.40}$ & $91.56_{\pm3.33}$ & $95.36_{\pm0.48}$ & $82.61_{\pm17.98}$ & $59.29_{\pm1.89}$ & 83.21 & $93.34_{\pm3.41}$ & $71.29_{\pm7.48}$ & $93.12_{\pm1.66}$ & $79.80_{\pm1.08}$ & $88.50_{\pm3.99}$ & 85.21 \\
    \rowcolor[rgb]{ .949,  .949,  .949} \textbf{FastGPT-Mult} & $\textbf{99.96}_{\pm0.03}$ & $\textbf{99.91}_{\pm0.06}$ & $\textbf{99.99}_{\pm0.01}$ & $\textbf{99.65}_{\pm0.29}$ & $\textbf{88.46}_{\pm2.30}$ & \textbf{97.59} & $\textbf{96.65}_{\pm1.66}$ & $\textbf{82.38}_{\pm3.91}$ & $\textbf{96.97}_{\pm0.40}$ & $\textbf{85.38}_{\pm0.61}$ & $\textbf{94.72}_{\pm1.00}$ & \textbf{91.22} \\
    \midrule
    Binoculars & $98.56_{\pm0.19}$ & $99.46_{\pm0.16}$ & $99.88_{\pm0.03}$ & $95.66_{\pm0.54}$ & $93.25_{\pm0.38}$ & 97.36 & $93.93_{\pm0.38}$ & $75.00_{\pm1.73}$ & $90.40_{\pm0.38}$ & $85.19_{\pm1.16}$ & $90.35_{\pm0.79}$ & 86.97 \\
    Binoculars-M & $99.36_{\pm0.08}$ & $99.85_{\pm0.05}$ & $99.70_{\pm0.11}$ & $96.39_{\pm0.36}$ & $\textbf{95.94}_{\pm0.51}$ & 98.25 & $86.82_{\pm3.20}$ & $74.48_{\pm1.98}$ & $77.21_{\pm3.29}$ & $75.15_{\pm2.72}$ & $80.92_{\pm2.12}$ & 78.92 \\
    \rowcolor[rgb]{ .949,  .949,  .949} \textbf{Binoculars-Mult} & $\textbf{99.82}_{\pm0.06}$ & $\textbf{99.88}_{\pm0.08}$ & $\textbf{99.92}_{\pm0.05}$ & $\textbf{99.12}_{\pm0.60}$ & $93.48_{\pm0.68}$ & \textbf{98.45} & $\textbf{98.34}_{\pm0.39}$ & $\textbf{87.35}_{\pm2.20}$ & $\textbf{97.38}_{\pm0.43}$ & $\textbf{86.88}_{\pm1.13}$ & $\textbf{96.50}_{\pm0.73}$ & \textbf{93.29} \\
    \midrule
    FourierGPT & $99.68_{\pm0.04}$ & $99.64_{\pm0.15}$ & $99.73_{\pm0.20}$ & $99.74_{\pm0.12}$ & $52.38_{\pm1.68}$ & 90.23 & $98.15_{\pm0.77}$ & $86.75_{\pm1.11}$ & $\textbf{97.79}_{\pm0.17}$ & $84.39_{\pm1.06}$ & $96.30_{\pm0.32}$ & 92.68 \\
    FourierGPT-M & $99.72_{\pm0.20}$ & $99.63_{\pm0.16}$ & $99.88_{\pm0.05}$ & $99.80_{\pm0.15}$ & $52.19_{\pm1.69}$ & 90.24 & $\textbf{98.38}_{\pm0.42}$ & $\textbf{86.86}_{\pm0.88}$ & $97.71_{\pm0.35}$ & $84.57_{\pm0.70}$ & $96.14_{\pm0.40}$ & 92.73 \\
    \rowcolor[rgb]{ .949,  .949,  .949} \textbf{FourierGPT-Mult} & $\textbf{99.99}_{\pm0.00}$ & $\textbf{99.97}_{\pm0.02}$ & $\textbf{99.99}_{\pm0.00}$ & $\textbf{99.94}_{\pm0.04}$ & $\textbf{64.41}_{\pm2.60}$ & \textbf{92.86} & $98.28_{\pm0.23}$ & $86.52_{\pm1.29}$ & $97.56_{\pm0.40}$ & $\textbf{85.46}_{\pm1.03}$ & $\textbf{96.48}_{\pm0.56}$ & \textbf{92.86} \\
    \midrule
    AdaGPT & $86.14_{\pm1.48}$ & $87.18_{\pm0.90}$ & $92.69_{\pm0.82}$ & $89.55_{\pm0.34}$ & $58.49_{\pm1.28}$ & 82.81 & $96.67_{\pm0.65}$ & $80.57_{\pm1.23}$ & $92.46_{\pm1.21}$ & $78.29_{\pm0.93}$ & $92.96_{\pm0.40}$ & 88.19 \\
    AdaGPT-M & $87.87_{\pm0.97}$ & $88.74_{\pm1.32}$ & $94.82_{\pm0.52}$ & $91.22_{\pm0.38}$ & $58.13_{\pm1.69}$ & 84.16 & $95.99_{\pm0.39}$ & $75.43_{\pm1.45}$ & $89.70_{\pm0.33}$ & $62.15_{\pm1.31}$ & $89.02_{\pm0.83}$ & 82.46 \\
    \rowcolor[rgb]{ .949,  .949,  .949} \textbf{AdaGPT-Mult} & $\textbf{99.98}_{\pm0.01}$ & $\textbf{99.91}_{\pm0.06}$ & $\textbf{99.99}_{\pm0.00}$ & $\textbf{99.67}_{\pm0.27}$ & $\textbf{86.94}_{\pm1.53}$ & \textbf{97.30} & $\textbf{98.05}_{\pm0.51}$ & $\textbf{85.60}_{\pm1.96}$ & $\textbf{96.92}_{\pm0.46}$ & $\textbf{82.24}_{\pm1.43}$ & $\textbf{95.30}_{\pm0.65}$ & \textbf{91.62} \\
    \midrule
    DetectLLM & $54.70_{\pm1.15}$ & $46.40_{\pm1.46}$ & $47.32_{\pm0.17}$ & $42.59_{\pm1.26}$ & $31.46_{\pm2.11}$ & 44.49 & $70.21_{\pm1.27}$ & $63.13_{\pm1.55}$ & $72.73_{\pm1.35}$ & $61.57_{\pm0.95}$ & $70.07_{\pm2.41}$ & 67.54 \\
    DetectLLM-M & $73.19_{\pm11.72}$ & $67.31_{\pm15.46}$ & $73.56_{\pm12.97}$ & $59.03_{\pm16.76}$ & $46.57_{\pm22.37}$ & 63.93 & $63.21_{\pm13.01}$ & $59.95_{\pm9.48}$ & $63.91_{\pm13.44}$ & $60.29_{\pm7.13}$ & $64.37_{\pm13.15}$ & 62.35 \\
    \rowcolor[rgb]{ .949,  .949,  .949} \textbf{DetectLLM-Mult} & $\textbf{99.93}_{\pm0.02}$ & $\textbf{99.70}_{\pm0.37}$ & $\textbf{99.95}_{\pm0.02}$ & $\textbf{98.97}_{\pm1.00}$ & $\textbf{86.31}_{\pm2.80}$ & \textbf{96.97} & $\textbf{97.79}_{\pm0.58}$ & $\textbf{86.55}_{\pm2.16}$ & $\textbf{97.17}_{\pm1.00}$ & $\textbf{84.59}_{\pm1.10}$ & $\textbf{95.43}_{\pm1.09}$ & \textbf{92.30} \\
    \bottomrule
    \end{tabular}%
    \end{adjustbox}
  \label{tab: main_auroc}%
\end{table*}%
\begin{figure*}[t]
	\centering
	\includegraphics[width=1.\linewidth]{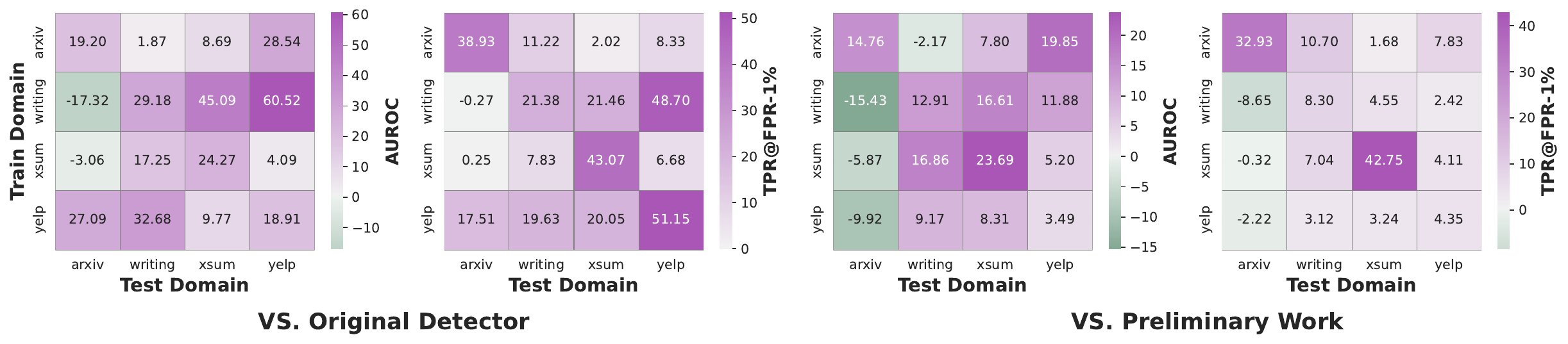}
 \vspace{-0.6cm}
	\caption{The performance improvement compared with original detector (left) and our preliminary work. Here the base detector is DetectLLM. Values greater than 0 indicate an enhanced effect.}
	\label{fig: cross_DetectLLM}
\end{figure*}

\subsubsection{Performance across Different LLMs}

Tables \ref{tab: main_tpr} and \ref{tab: main_auroc} report the cross-LLM results under TPR@FPR-1\% and AUROC when detectors are trained on GPT4 texts. More results about other LLM text training and other datasets can be found in Section V-C of the Supplementary Material. While our preliminary work (\textbf{-M}) already improves most base detectors, the proposed method (\textbf{-Mult}) further delivers stronger cross-LLM generalization, especially under the stricter TPR@FPR-1\% metric. For example, on Essay, the average TPR@FPR-1\% improves from 52.41\% to 77.87\% to 82.65\% for Likelihood, and from 0.15\% to 37.17\% to 82.32\% for DetectGPT; similar gains are also observed for other detectors. On TruthfulQA, which consists of short texts, our advantage is also evident for most detectors; e.g., Likelihood improves from 46.38\% to 50.82\% to 69.30\%. 
The same pattern is reflected by AUROC. Compared with TPR@FPR-1\%, the absolute gains in AUROC are generally smaller, which is expected because several baselines already achieve relatively high AUROC scores. Even so, the proposed method still provides consistent improvements.
Overall, extensive cross-LLM experiments demonstrate the effectiveness of our multi-level contextual relation modeling in capturing MGT features.

\subsubsection{Performance across Different Domains}

We further evaluate cross-domain generalization on the DetectRL benchmark, which comprises four high-risk domains: \emph{arXiv}, \emph{Writing Prompts}, \emph{XSum}, and \emph{Yelp Reviews}. Each detector is trained on one domain and tested on the remaining domains, and Fig. \ref{fig: cross_DetectLLM} summarizes the performance gains of the DetectLLM detector (additional enhanced results for more detectors are provided in Section V-D of the Supplementary Material). It is evident that the proposed framework improves cross-domain detection in most settings, indicating that contextual modeling is not limited to a specific content domain and transfers well across heterogeneous writing styles and topics. For example, the average AUROC gain compared with the original detector across the 16 train--test pairs is about \(+19.1\), and the average AUROC gain compared with our preliminary work is about \(+7.2\). We attribute this out-of-domain generalization to the fact that, building upon preliminary local relation modeling, the proposed rule-support reasoning module further leverages global contextual relations, making the enhanced detector less dependent on superficial domain cues and therefore more transferable across domains. 

\begin{figure*}[t]
	\centering
	\includegraphics[width=1.\linewidth]{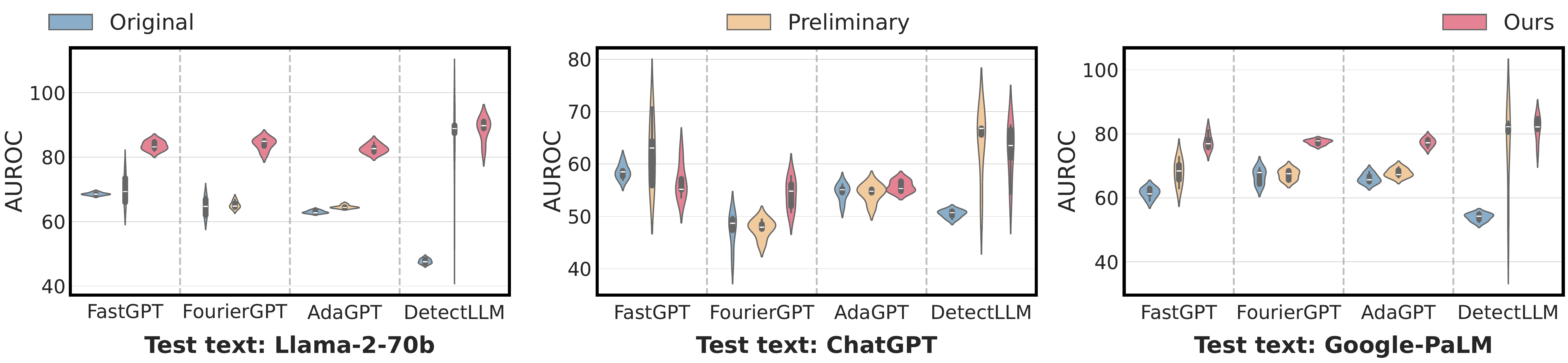}
 \vspace{-0.6cm}
	\caption{Detection performance concerning AUROC under different mixed texts. All detectors are trained on pure Llama-2-70b texts.}
	\label{fig: mix}
\end{figure*}

\begin{figure*}[t]
	\centering
	\includegraphics[width=1.\linewidth]{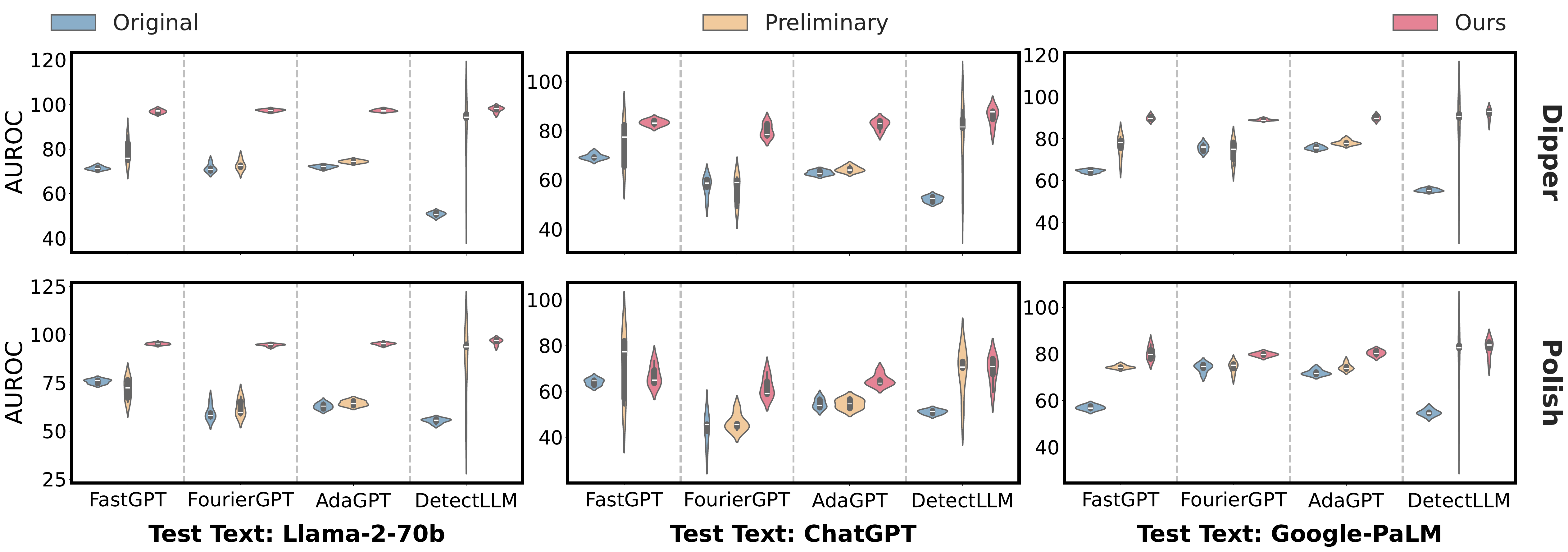}
 \vspace{-0.6cm}
	\caption{Detection performance under Dipper and Polish paraphrasing texts. All detectors are trained on Llama-2-70b texts.}
	\label{fig: paraphrasing}
\end{figure*}

\subsubsection{Performance against Mixed Texts}
In practice, human--AI collaboration is pervasive, so a detector must distinguish not only pure MGT from HGT, but also mixed texts in which machine-generated content is partially blended with human-written sentences. We therefore evaluate the proposed framework on the mixed-text setting of DetectRL, where 1/4 of the machine-generated sentences are replaced by human-written ones while the label remains MGT. The results are shown in Fig. \ref{fig: mix}. The proposed method improves AUROC over both the original detector and our preliminary work in most settings. This indicates that the proposed multi-level design is effective even when the machine signal is partially diluted by human-written content. Compared with our preliminary work, our additional gains suggest that global contextual relation modeling is helpful in mixed-text scenarios, since mixed texts are harder to distinguish using only local score smoothing and require more global structural cues.
Undeniably, the proposed method performs poorly in detecting ChatGPT text. This is because its performance in this specific context approaches that of random chance (e.g., an AUROC close to 0.5, and TPR@FPR-1\%$<0.1$), indicating that the detection scores generated by the original detector are completely unreliable, which in turn leads to suboptimal performance in the enhanced one. However, as demonstrated by the detection of Llama-2-70b and Google-PaLM texts, the proposed method proves effective provided that the underlying detection scores have a certain degree of predictive value. We emphasize that our efforts toward enhancement should focus on developing effective detectors rather than weak, nearly stochastic ones, as the latter lack practical utility.

\subsubsection{Performance against Paraphrasing Attacks}
Prior work \cite{sadasivan2023can} has shown that MGT detectors are particularly vulnerable to paraphrasing attacks, because paraphrasing can preserve the original semantics while concealing surface-level MGT patterns. We therefore evaluate the robustness of the proposed method on two widely used paraphrasing attacks in DetectRL, namely \emph{Polish} and \emph{Dipper}, where the detector is trained on clean texts from Llama-2-70b and tested on paraphrased texts from different source LLMs. The results w.r.t. AUROC are shown in Fig. \ref{fig: paraphrasing}. The proposed method consistently outperforms both the original detector and our preliminary work in most settings. This result indicates that the proposed multi-level framework remains effective even when the original MGT has been substantially rewritten. In particular, the gains over the preliminary work suggest that global contextual relation modeling provides additional robustness beyond local calibration alone, since paraphrasing not only introduces local token-level perturbations but may also alter broader sequence-level score patterns. 

\begin{figure*}[t]
	\centering
	\includegraphics[width=1.\linewidth]{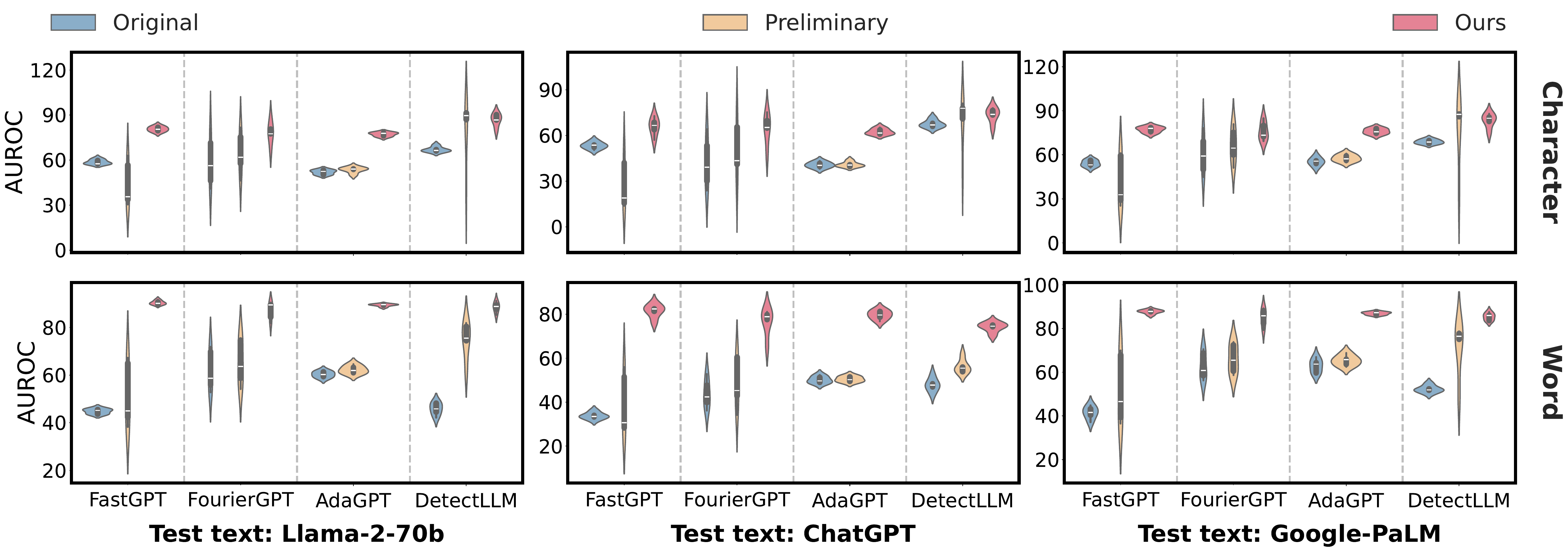}
 \vspace{-0.6cm}
	\caption{Detection performance concerning AUROC under adversarial texts concerning character perturbation. All detectors are trained on Llama-2-70b texts.}
	\label{fig: adversarial}
\end{figure*}

\begin{figure*}[t]
	\centering
	\includegraphics[width=1.\linewidth]{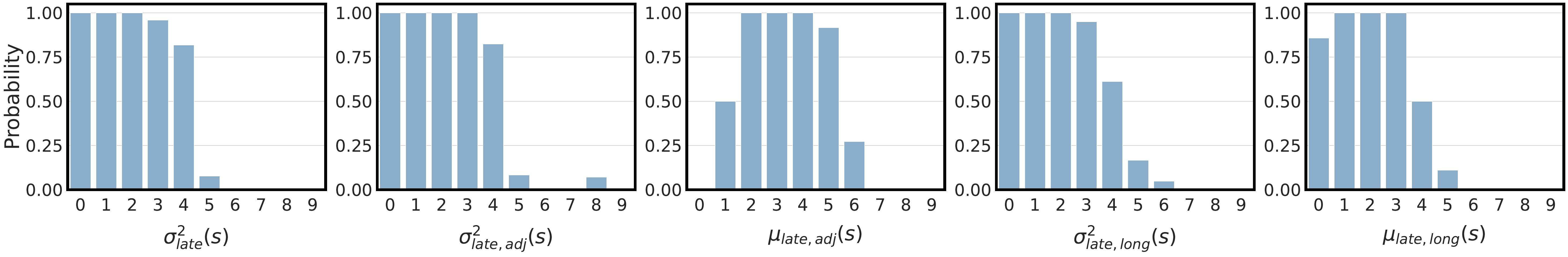}
 \vspace{-0.6cm}
	\caption{The empirical MGT probability associated with different threshold atoms induced by the global statistics. DetectLLM scores are used here.}
	\label{fig: rule_DetectLLM}
\end{figure*}

\subsubsection{Performance against Adversarial Attacks}
Beyond paraphrasing, MGT detection is also vulnerable to adversarial attacks \cite{zhu2025slack}. Therefore, we further evaluate robustness to stronger adversarial perturbations at \emph{character}, and \emph{word} granularities. The corresponding comparison results are shown in Fig. \ref{fig: adversarial}. A highly consistent trend can be observed: across nearly all settings, the proposed method outperforms both the original detector and our preliminary local-only version, indicating that the proposed multi-level design substantially improves adversarial robustness. A plausible explanation is that such attacks act as structured noise on detector scores: the local calibration module suppresses token-level corruption by exploiting short-range consistency, while the global rule-support reasoning module further recovers more stable text-level cues from score statistics. Their combination therefore provides stronger robustness than either the original detector or the preliminary local-only framework. 

\subsection{Rule Visualization}

To provide an intuitive understanding of the effectiveness of the proposed rule-support reasoning module, Fig. \ref{fig: rule_DetectLLM} visualizes the empirical MGT probability associated with different rule patterns induced by the global statistics of DetectLLM. More results can be found in Section V-E of the Supplementary Material. A clear trend can be observed: rules corresponding to stronger latter-part stability are consistently assigned higher MGT probability, whereas rules associated with larger score fluctuations tend to be less indicative of MGT. This observation is highly consistent with the distributional evidence in Fig. \ref{fig: motivation_position_DetectLLM_Essay}, where MGT is concentrated in the low-variance and low-difference regions, while HGT exhibits a broader and more dispersed distribution. Therefore, Fig. \ref{fig: rule_DetectLLM} provides a qualitative validation of the rule design: the rule-support reasoning module transforms empirically supported global regularities into explicit rules whose support scores remain aligned with the actual class tendency in the data. This result further justifies incorporating rule-support reasoning into our framework, as it captures text-level cues that complement local calibration and are genuinely informative for distinguishing MGT from HGT. 

\subsection{Comparison with Probabilistic Rule-based Method}
\label{sec: ablation_probability}
\begin{figure*}[t]
	\centering
	\includegraphics[width=1.\linewidth]{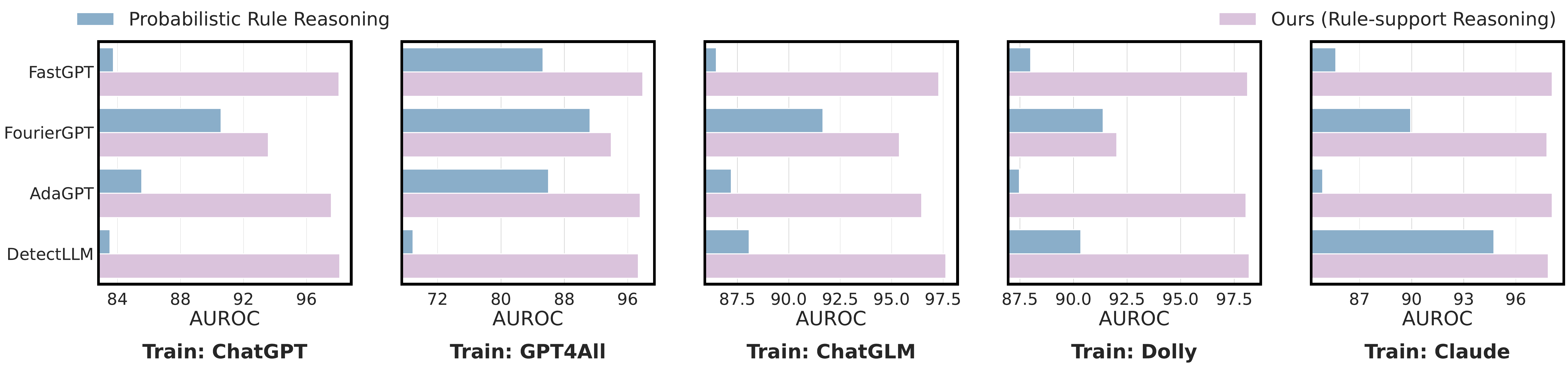}
 \vspace{-0.6cm}
	\caption{Performance comparison with probabilistic rule-based method on the Essay dataset. The reported results are the average performance across all LLM-generated texts.}
	\label{fig: ablation_probability_Essay_auc}
\end{figure*}
\begin{figure*}[t]
	\centering
	\includegraphics[width=1.\linewidth]{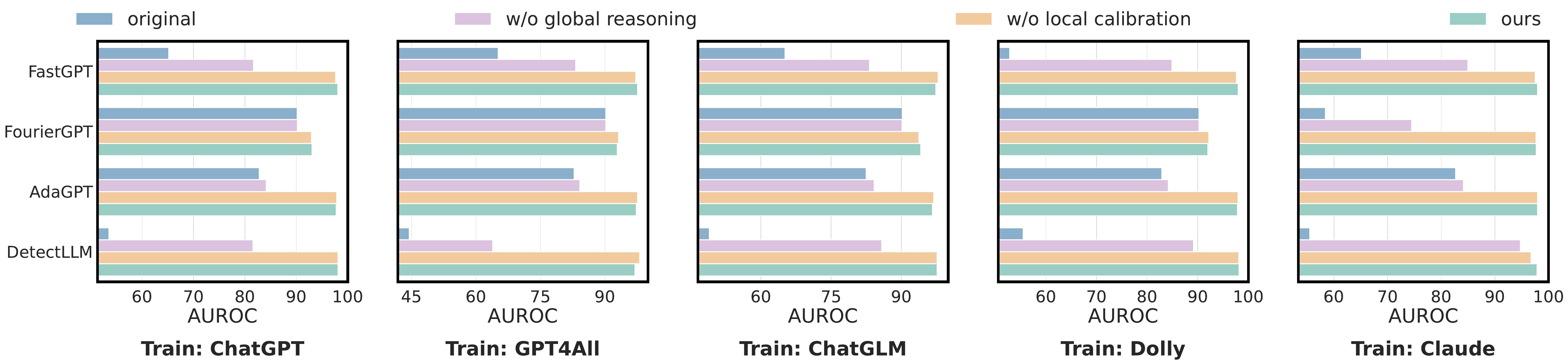}
 \vspace{-0.6cm}
	\caption{Ablation results concerning AUROC on the Essay dataset. The reported results are the average performance across all LLM-generated texts.}
	\label{fig: ablation_Essay_auc}
\end{figure*}

We further compare the proposed rule-support reasoning with a more direct probabilistic rule-based alternative that estimates the detection score from the full conjunction rule. As shown in Fig. \ref{fig: ablation_probability_Essay_auc}, our rule-support reasoning consistently achieves better AUROC across different training sources and base detectors. This result is consistent with the motivation in Section V-B. Although the probabilistic formulation is ideal in principle, directly estimating the conditional probability of a full rule is unreliable in our setting, because the training set is relatively small while the number of possible conjunction rules grows combinatorially after discretization. As a result, many full rules are supported by very few samples, making the resulting estimates sparse and unstable.

\subsection{Ablation Study}
We introduce local calibration and global reasoning modules to model multi-level relations among contextual tokens. In this section, we verify their effectiveness through ablation studies, denoted as "w/o global" and 'w/o local'. Results on the Essay dataset are shown in Fig. \ref{fig: ablation_Essay_auc}, with additional results available in Section V-F of the Supplementary Material. Since the preliminary work has already established the effectiveness of local calibration, we focus here on the additional value brought by the proposed global rule-support reasoning module.
It is evident that the full model consistently achieves the best performance, whereas removing the global reasoning branch leads to a noticeable drop across almost every setting, especially for challenging detectors such as FastGPT and DetectLLM. This directly verifies that our gains are not merely inherited from the preliminary local module, but are substantially supported by the newly introduced global reasoning component. Moreover, the variant without local calibration still remains competitive in many settings, indicating that the global rule-support reasoning module itself already contributes strong discriminative power.

\begin{figure*}[t]
	\centering
	\includegraphics[width=1.\linewidth]{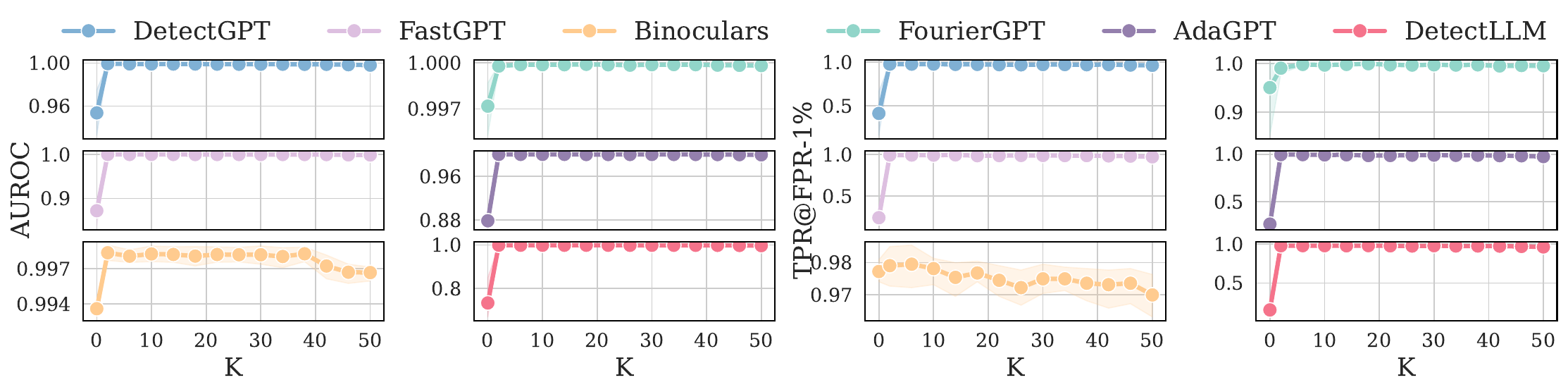}
 \vspace{-0.6cm}
	\caption{Detection performance at different numbers of rules on the Essay dataset. All detectors are trained on ChatGPT texts.}
	\label{fig: sensitive_k_Essay}
\end{figure*}

\begin{figure*}[t]
	\centering
	\includegraphics[width=1.\linewidth]{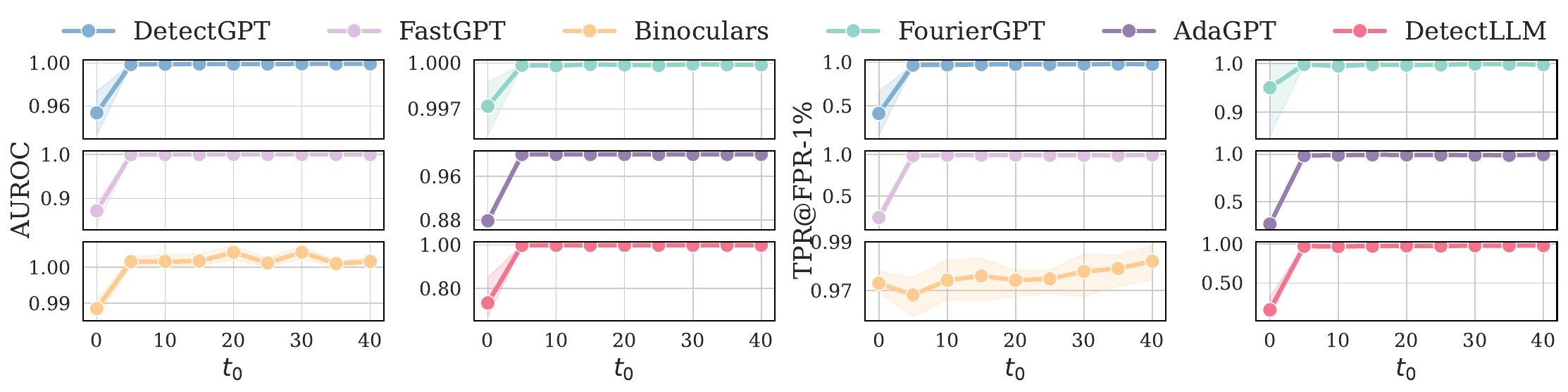}
 \vspace{-0.6cm}
	\caption{Detection performance at different initial part lengths on the Essay dataset. All detectors are trained on ChatGPT texts.}
	\label{fig: sensitive_t_Essay}
\end{figure*}

\subsection{Sensitivity Analysis}
Since the preliminary work has already provided a sensitivity analysis for the local calibration module, we here focus on the key hyperparameter introduced by the global rule-support reasoning module.

\textbf{Bucket Number $K$}, which is related to the number of rules. As shown in Fig. \ref{fig: sensitive_k_Essay} (more results are in Section V-G of the Supplementary Material), the proposed method remains relatively stable over a broad range of $K$ values, indicating that the global reasoning branch is not overly sensitive to this discretization choice and thus enjoys good practical robustness. Undeniably, when $K$ becomes too large, performance may tend to decline. This is because finer discretization partitions the statistic space into too many small buckets, so that each bucket contains fewer training samples. This weakens the reliability of the corresponding atom-level support estimates and makes the resulting rule-support scores more sparse and unstable. Based on this trade-off, we use a moderate default setting, i.e., $K=10$, in all experiments. 

\textbf{Initial Part Length $t_0$}, which determines the starting position of the latter part used in the global statistics. As shown in Fig. \ref{fig: sensitive_t_Essay}, the proposed method remains stable over a broad range of $t_0$ values, indicating that the global rule-support reasoning module is not overly sensitive to the exact choice of the latter-part starting point. Even with a very small value of $t_0$, the detection performance remains outstanding when compared to not utilizing global rule-support reasoning ($t_0=0$). This highlights the practicality of our method, since it does not require extensive parameter tuning.

\subsection{Running Time}

\begin{table}[t]
  \centering
  \caption{Running time (s) of training and inference phases.}
  \renewcommand\arraystretch{0.9}   \begin{adjustbox}{width=0.45\textwidth}   \setlength{\tabcolsep}{0.008\linewidth}
    \begin{tabular}{c|cc|cc}
    \toprule
    \multirow{2}[0]{*}{Method} & \multicolumn{2}{c|}{Train} & \multicolumn{2}{c}{Inference} \\
    \noalign{\smallskip} \cline{2-5}\noalign{\smallskip} 
          & Essay & TruthfulQA & Essay & TruthfulQA \\
          \midrule
    DetectGPT & 204.61  & 385.24  & 925.40  & 1738.72  \\
    DetectGPT-M & 206.45  & 389.13  & 932.41  & 1754.73  \\
    \textbf{DetectGPT-Mult} & 207.06  & 390.11  & 934.72  & 1765.17  \\
    \midrule
    FastGPT & 60.98  & 45.23  & 283.85  & 206.16  \\
    FastGPT-M & 63.11  & 47.07  & 287.57  & 217.30  \\
    \textbf{FastGPT-Mult} & 63.96  & 48.07  & 294.93  & 224.03  \\
    \midrule
    Binoculars & 44.25  & 36.24  & 200.98  & 167.47  \\
    Binoculars-M & 45.77  & 37.87  & 207.27  & 173.25  \\
    \textbf{Binoculars-Mult} & 46.94  & 38.80  & 214.47  & 181.54  \\
    \midrule
    FourierGPT & 20.34  & 15.03  & 97.18  & 75.12  \\
    FourierGPT-M & 21.70  & 16.39  & 98.62  & 79.79  \\
    \textbf{FourierGPT-Mult} & 22.81  & 17.26  & 111.01  & 85.33  \\
    \midrule
    AdaDetectGPT & 61.49  & 42.69  & 285.84  & 197.96  \\
    AdaDetectGPT-M & 63.27  & 44.34  & 286.38  & 199.76  \\
    \textbf{AdaDetectGPT-Mult} & 64.44  & 44.95  & 293.75  & 205.39  \\
    \midrule
    DetectLLM & 37.27  & 41.79  & 169.12  & 189.79  \\
    DetectLLM-M & 38.90  & 43.70  & 176.15  & 200.97  \\
    \textbf{DetectLLM-Mult} & 39.57  & 44.94  & 178.87  & 202.48  \\
    \bottomrule
    \end{tabular}%
    \end{adjustbox}
  \label{tab: running_time}%
\end{table}%

Table \ref{tab: running_time} reports the training and inference time on four datasets. Overall, the proposed method introduces only a negligible runtime overhead, while providing substantially stronger detection performance. This trend is particularly clear for detectors whose dominant cost already lies in score computation, such as DetectGPT. These results are consistent with our design: the local calibration mainly involves sparse-dense operations over adjacent positions, while the global rule-support reasoning branch only computes a small number of low-dimensional statistics and bucket-based scores. Therefore, our method preserves the practical efficiency advantage and achieves improved performance with only negligible-to-moderate additional computational cost. This observation is also consistent with the complexity analysis in Section \ref{sec: joing_inference}, where the time complexity is only $O(NT+NM+MK)$.

\section{Conclusion}

In this paper, we have proposed a multi-level contextual token relation modeling framework for machine-generated text detection. By revisiting representative metric-based detectors under a unified view, we have identified a shared limitation: token-level detection scores can be biased by the stochasticity of LLM generation, while direct aggregation cannot explicitly correct such imprecision. To this end, we have modeled contextual relations among token-level scores from both local and global perspectives. Locally, we have captured Neighbor Similarity and Initial Instability through a lightweight Markov-informed calibration module. Globally, we have characterized score stability patterns of MGTs and introduced a rule-support reasoning module to model them explicitly. The two modules are integrated into a joint multi-level inference framework, leading to improved detection performance across multiple datasets and practical scenarios with low computational overhead. Nevertheless, the proposed framework is mainly applicable to metric-based detectors whose scores can be decomposed at the token level. In addition, future adaptive attacks may deliberately disrupt score stability patterns, which remains an important direction for further study.

\bibliographystyle{IEEEtran}
\bibliography{ref}


\end{document}